\documentclass[10pt,twocolumn,letterpaper]{article}

\usepackage{iccv}              

\usepackage{amsmath,amsfonts,bm}

\def\eqref#1{equation~\ref{#1}}

\def\1{\bm{1}}

\DeclareMathAlphabet{\mathsfit}{\encodingdefault}{\sfdefault}{m}{sl}
\SetMathAlphabet{\mathsfit}{bold}{\encodingdefault}{\sfdefault}{bx}{n}

\newcommand{\q}{\boldsymbol{r}}
\newcommand{\s}{\boldsymbol{s}}
\newcommand{\x}{\boldsymbol{x}}
\newcommand{\fs}{\boldsymbol{z}}
\newcommand{\Dt}{\Delta t}
\newcommand{\fdefo}{f_{defo}}
\newcommand{\ftrd}{f_{trd}}

\newcommand{\R}{\mathcal{R}}

\definecolor{iccvblue}{rgb}{0.21,0.49,0.74}
\usepackage[pagebackref,breaklinks,colorlinks,allcolors=iccvblue]{hyperref}
\usepackage{comment}
\usepackage{graphicx}
\usepackage{multirow} 
\usepackage{enumitem}
\usepackage{amssymb}
\usepackage{wrapfig}
\usepackage{pifont}
\usepackage{algpseudocode}
\usepackage{algorithm}

\usepackage{xcolor}
\usepackage{color,soul}

\usepackage[normalem]{ulem}

\newcommand{\Y}{\ding{51}}
\newcommand{\N}{\ding{55}}

\newcommand{\nickname}{TRACE}

\newcommand{\myrightleftarrows}[1]{\mathrel{\substack{\xrightarrow{#1} \\[-.9ex] \xleftarrow{#1}}}}

\def\paperID{8898} 
\def\confName{ICCV}
\def\confYear{2025}

\title{\nickname{}: Learning 3D Gaussian Physical Dynamics from Multi-view Videos}

\author{Jinxi Li, \quad Ziyang Song, \quad Bo Yang\footnotemark[1]\\
vLAR Group, The Hong Kong Polytechnic University\\
{\tt\small \{jinxi.li, ziyang.song\}@connect.polyu.hk, bo.yang@polyu.edu.hk}}

\begin{document}

\twocolumn[{%
\renewcommand\twocolumn[1][]{#1}%
    \maketitle
    \begin{center}
        \vspace{-15pt}
        \centering
        \includegraphics[width=1.0\linewidth]{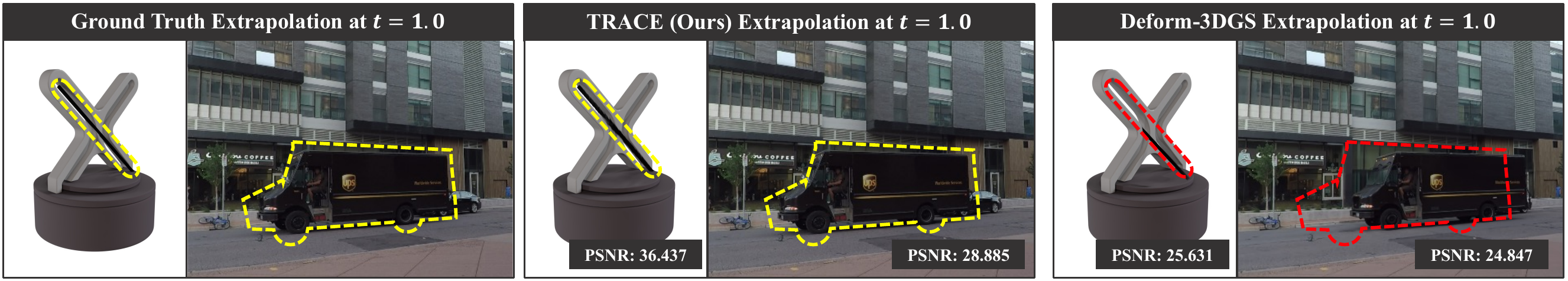}
        \vspace{-15pt}
        \captionof{figure}{Given video frames of a real-world dynamic scene, our \nickname{} can learn the underlying physics and  accurately predict the future motion of a rode passing through a hyperbolic slot and a track keep going forward, while the existing method cannot.}
        \label{fig:opening}
        \vspace{15pt}
    \end{center}
}]

\renewcommand{\thefootnote}{\fnsymbol{footnote}}
\footnotetext[1]{Corresponding Author}

\begin{abstract}
In this paper, we aim to model 3D scene geometry, appearance, and physical information just from dynamic multi-view videos in the absence of any human labels. By leveraging physics-informed losses as soft constraints or integrating simple physics models into neural nets, existing works often fail to learn complex motion physics, or doing so requires additional labels such as object types or masks. We propose a new framework named \textbf{\nickname{}} to model the motion physics of complex dynamic 3D scenes. The key novelty of our method is that, by formulating each 3D point as a rigid particle with size and orientation in space, we directly learn a translation rotation dynamics system for each particle, explicitly estimating a complete set of physical parameters to govern the particle's motion over time. Extensive experiments on three existing dynamic datasets and one newly created challenging synthetic datasets demonstrate the extraordinary performance of our method over baselines in the task of future frame extrapolation. A nice property of our framework is that multiple objects or parts can be easily segmented just by clustering the learned physical parameters. Our datasets and code are available at {\small\url{https://github.com/vLAR-group/TRACE}}.   
\end{abstract}   

\vspace{-0.35cm}
\section{Introduction}
\label{sec:intro}

Regarding our daily dynamic 3D scenes such as falling balls, rotating fans, and folding chairs, precisely modeling their geometry, appearance, and physical properties, and further predicting their future states are crucial for emerging applications in robotics, mixed reality, and embodied AI. With the advancement of recent 3D representations such as NeRF \citep{Mildenhall2020} and 3DGS \citep{Kerbl2023}, a plethora of works \citep{Pumarola2021,Yang2024,Wu2024} have been proposed to model various dynamic 3D scenes, achieving excellent performance in interpolating novel views within the observed time. However, they often fail to extrapolate future frames, essentially because they do not learn the underlying physics priors of 3D scenes.     

To learn physics priors, existing methods mainly consist of two categories: 1) physics-informed neural network (PINN) based methods \citep{Raissi2019} which integrate the governing partial differential equations (PDEs) into loss functions to drive neural networks to learn physically plausible dynamic 3D scenes such as floating smoke \citep{Qiu2024} and simple moving objects \citep{Li2023}. While demonstrating promising results in modeling 3D geometry and physics such as velocity and viscosity, these methods usually need boundary constraints such as accurate object/foreground masks which may not always be available in practice. In addition, adding PINN losses is not a free lunch, but significantly sacrifices the efficiency in training and accuracy at boundary regions. 2) Physics model based methods \citep{Jonathan2020,Zhong2024,Whitney2024} which encode various physics systems into neural networks to model elastic objects, fluids, \etc{}. Thanks to the explicit physics priors added, these methods obtain impressive results in physical properties learning and simulation. Nevertheless, they are often limited to specific types of objects, materials, or motions due to the lack of generality of encoded physics priors, thus being unable to predict future motions of complex dynamic 3D objects and scenes. 

In this paper, we aim to introduce a new framework to model dynamic 3D scenes just from multi-view RGB videos, without needing any additional human labels such as object types or masks, ultimately being able to predict future frames viewing from arbitrary angles. Among various physical properties of a dynamic 3D scene, following the recent work NVFi \citep{Li2023c}, we also choose to learn a velocity field as it directly governs 3D scene movement. However, to accurately learn the physical velocity from RGB videos is extremely challenging, essentially due to the lack of sufficient physics constraints from raw color pixels. This problem is even harder when multiple objects or parts are undergoing rather different motion patterns. For example, regarding two adjacent objects moving in opposite directions in 3D space, the velocity of neighboring 3D surface points at the intersection region tends to have particularly distinct patterns. This means that the latent representation of per-point dynamics in 3D space could be discrete in nature. Therefore, it is more desirable to model per-point dynamics independently, thus every point having its unique motion. For generality, we regard each 3D point in space as a rigid particle with its size and orientation. If its size is zero, the rigid particle degenerates to a point.    

With this insight, for each rigid particle in space, we propose to learn an independent dynamics system that includes a complete set of physical parameters to govern its motion over time. According to the laws of classical mechanics, for a specific rigid particle traversing 3D space over time, its motion can always be regarded as a rotational movement about a rotation center which has its own translation. In this regard, we choose to learn a translation rotation dynamics system for each rigid particle, allowing its future motion to be derived accordingly. Alongside learning the core dynamics, we must also model the geometry and appearance of 3D scenes. In this paper, we naturally choose 3D Gaussian Splatting (3DGS) \citep{Kerbl2023} as the representation, thanks to its unprecedented fidelity in reconstruction and its particle (a Gaussian kernel) based representation in nature, which shares the basic concept of our defined rigid particle. 

Our framework consists of two major components: 1) a \textbf{3D scene representation module} to learn dynamic scene geometry and appearance at a canonical timestamp, which is implemented by a vanilla 3DGS \citep{Kerbl2023}, though other variants can be adopted as well; 2) a \textbf{translation rotation dynamics system module} to learn a full set of physical parameters for each input rigid particle, which is just realized by multilayer perceptrons (MLPs). Based on these system parameters, the rigid particle's velocity is then derived according to the laws of classical mechanics, without needing additional physics priors such as PINN \citep{Raissi2019} in training. 

The key to our framework is the second module which simply regards each 3D Gaussian kernel as a rigid particle and takes it as input into MLPs. Nevertheless, we empirically find that it is hard to optimize this module due to the inaccuracy and instability of Gaussian kernels regressed at early training epochs. To tackle this issue, we simply train an auxiliary deformation field in parallel with our second module using an existing work such as \citep{Yang2024} and \citep{Wu2024}.

Different from current works for modeling dynamic scenes, including NeRF-based methods, \eg{}, D-NeRF\citep{Pumarola2021}/ TiNeuVox\citep{Fang2022}/ HexPlane\citep{Cao2023a}, and 3DGS-based methods such as DefGS \citep{Yang2024} and 4DGS \citep{Wu2024}, our core novelty is the introduced translation rotation dynamics system with an effective optimization strategy, allowing us to truly learn physical parameters, ultimately achieving future frame extrapolation. By comparison, all those existing methods fail to do so, as extensively verified in Tables \ref{tab:exp_extrapolation_only}.

By leveraging 3D Gaussians to model scene geometry and appearance, our method, named \textbf{\nickname{}}, learns \textbf{t}ranslation \textbf{r}otation dyn\textbf{a}mi\textbf{c}s syst\textbf{e}ms for estimating accurate physical dynamics. Figure \ref{fig:opening} shows our qualitative results. Our contributions are:
\begin{itemize}[leftmargin=*] 
\setlength{\itemsep}{1pt}
\setlength{\parsep}{1pt}
\setlength{\parskip}{1pt}
    \item We introduce a new framework to model motion physics of complex dynamic 3D scenes, without needing prior knowledge of object shapes, types, or masks.  
    \item We propose to learn a translation rotation dynamics system for each 3D rigid particle, thus allowing the velocity field to be derived without needing additional physics constraints in training.    
    \item We demonstrate superior results in future frame extrapolation on three existing datasets, and one newly collected synthetic dataset with extremely challenging dynamics. 
\end{itemize}

The concurrent work FreeGave \cite{Li2025} addresses a similar task to ours. However, the key difference is that \nickname{} explicitly models the changes (accelerations, jerks, or higher order) of physical motions, allowing continuous derivation of velocities over time, whereas FreeGave simply fits a velocity network over observed time in an implicit manner.

\section{Related Works}
\label{sec:literature}

\textbf{3D Shape Representations}:
Static 3D objects and scenes are traditionally represented by voxels, point clouds, meshes, \etc{}, but they usually have limited representation capabilities due to the nature of discretization. Recently, implicit representations have been developed in the literature, including occupancy fields (OF) \citep{Mescheder2019,Chen2019g}, un/signed distance fields (U/SDF) \citep{Park2019,Chibane2020a}, and radiance fields (NeRF) \citep{Mildenhall2020}. Although demonstrating excellent performance in novel view synthesis and shape reconstruction, they are time-consuming to render 2D images or extract 3D shapes due to the integration of their continuous coordinate-based representations. To tackle this issue, the very recent 3D Gaussian Splatting \citep{Kerbl2023} turns to represent a 3D shape as a set of explicit Gaussian kernels with various properties, achieving real-time rendering speed. In our framework, we adopt this particle-based representation, as it is amenable to our particle-based physics learning framework. 

\textbf{Dynamic 3D Reconstruction}: Recent advances in dynamic 3D reconstruction primarily follow the development of static 3D techniques such as SDF, NeRF, and Gaussian Splatting. To model the temporal relationship, existing works \citep{Tretschk2021,Li2021c,Barron2021a,Gao2021,Tian2023,Liu2023,Cao2023a,Fridovich-Keil2023,Cai2022,Fang2022,Li2022,Park2021,You2023,Du2021a,Park2023,Xian2021,Wang2021i,Liu2024} usually add the time dimension into static 3D representations to learn a motion or deformation field for rigid or deformable objects and scenes. Despite achieving excellent performance in novel view synthesis, especially when integrating 3DGS as the backbone \citep{Wu2024,Yang2024,Li2024,Lei2024,Lin2024,Lu2024}, these works can only interpolate 2D views within the observed time, instead of predicting physically meaningful future frames. Basically, this is because the commonly learned motion or deformation field does not encode physics priors in nature, but just fits the correlation between pixels. In this paper, the key difference between these works and ours is that we separately learn translation rotation dynamics systems for 3D rigid particles, thus allowing us to infer physically meaningful future frames, but they cannot.

\textbf{3D Physics Learning}: To learn various physical properties for 3D objects and scenes, the recent physics-informed neural networks (PINN) \citep{Raissi2019,Mishra2023,Raissi2020,Hao2023,Baieri2023,Chalapathi2024,Qiu2024,Zhao2024} are widely applied to convert PDEs into loss functions as soft constraints, driving neural networks to learn physically meaningful targets. However, it is often inefficient to train PINNs due to the large amount of data samples needed to regularize, and the soft constraints are usually not sufficient to obtain satisfactory results. In this paper, we do not rely on such inefficient PINN losses to incorporate physics priors to train neural networks.
Another line of works \citep{Qiao2022,Deng2023,Xue2023,Franz2023,Whitney2024} integrate explicit physics systems such as springs, graphs, \etc{}, into the learning  process to model elastic objects \citep{Zhong2024,Zhang2024,Liu2024a}, fluids \citep{Jonathan2020,Lienen2024}, \etc{}, achieving impressive results in physics learning and simulation. In this paper, we also opt to learn physics systems. However, the core difference is that we learn a translation rotation dynamics system which is applicable to common deformation and transformation dynamics, whereas existing works often learn a spring or fluid system only applicable to elastic objects or fluids.    

\section{\nickname{}}
\label{sec:method}

Our framework mainly comprises two modules together with an auxiliary deformation field to model 3D geometry, appearance, and physics. Given dynamic multi-view RGB videos with known camera poses and intrinsics, the 3D scene representation module aims to learn a set of 3D Gaussian kernels to represent the 3D scene geometry and appearance in a canonical space. The auxiliary deformation field is designed to predict the translation and distortion of each Gaussian kernel given the current training time $t$. For these two components, we simply follow the design of existing works \citep{Kerbl2023,Yang2024}
briefly elaborated in Section \ref{sec:preliminary}. Notably, the deformation field alone cannot extrapolate frames beyond the training time. Our core module of the translation rotation dynamics system aims to learn a set of physical parameters for each 3D rigid particle, governing its motion dynamics over time, which is detailed in Section \ref{sec:transRot}.      

\subsection{Preliminary}\label{sec:preliminary}

For the input multi-view RGB videos, $T$ represents the greatest timestamp in training and $N$ the total number of cameras. For training stability, we first use all frames $\{\boldsymbol{I}^1_0 \cdots \boldsymbol{I}^n_0 \cdots \boldsymbol{I}^N_0\}$ at time $t=0$ to train a reasonable static 3DGS model as an initialization of the 3D scene geometry and appearance, and then use the remaining frames to jointly optimize our translation and rotation dynamics system and the auxiliary deformation field. 

\textbf{Canonical 3D Gaussians}: Following the vanilla 3DGS \citep{Kerbl2023}, we employ a set of learnable 3D Gaussian kernels $\boldsymbol{G}_0$ to represent the canonical scene geometry and appearance at $t=0$. Each kernel is parameterized by a 3D position $\boldsymbol{x}_0$, covariance matrix obtained from quaternion $\boldsymbol{r}_0$, scaling $\boldsymbol{s}_0$, opacity $\sigma$, and color $\boldsymbol{c}$ computed from spherical harmonics. Following prior works \citep{Yang2024,Wu2024}, we assume the opacity $\sigma$ and color $\boldsymbol{c}$ of each Gaussian will not be updated, but constantly associated with the kernel and transported over time.  

Given the $N$ images at timestamp $t=0$, we either initialize all canonical 3D Gaussian kernels $\boldsymbol{G}_0$ randomly or based on sparse points created by SfM \citep{Schonberger2016}. To train all kernels, we exactly follow 3DGS \citep{Kerbl2023} by 1) projecting Gaussian kernels into camera space, 2) rendering the projected kernels into image space, and 3) optimizing all kernel parameters via $\ell_1$ and $\ell_{ssim}$ losses used in 3DGS as follows.
\begin{equation}
\setlength{\abovedisplayskip}{3pt}
\setlength{\belowdisplayskip}{3pt}
    \underbrace{ \bigl\{ \cdot \cdot (  \boldsymbol{x}_0, \boldsymbol{r}_0, \boldsymbol{s}_0, \sigma, \boldsymbol{c})  \cdot \cdot \bigr\} }_{\boldsymbol{G}_0}  
    \mathrel{\mathop{\myrightleftarrows{\rule{1.6cm}{0cm}}}^{\mathrm{project+render}}_{\mathrm{\ell_1 + \ell_{ssim}}}}
    \bigl\{\boldsymbol{I}^1_0  \cdot \cdot \boldsymbol{I}^N_0 \bigr\}
\end{equation}

\begin{figure*}[t]
    \centering
\includegraphics[width=1.0\linewidth]{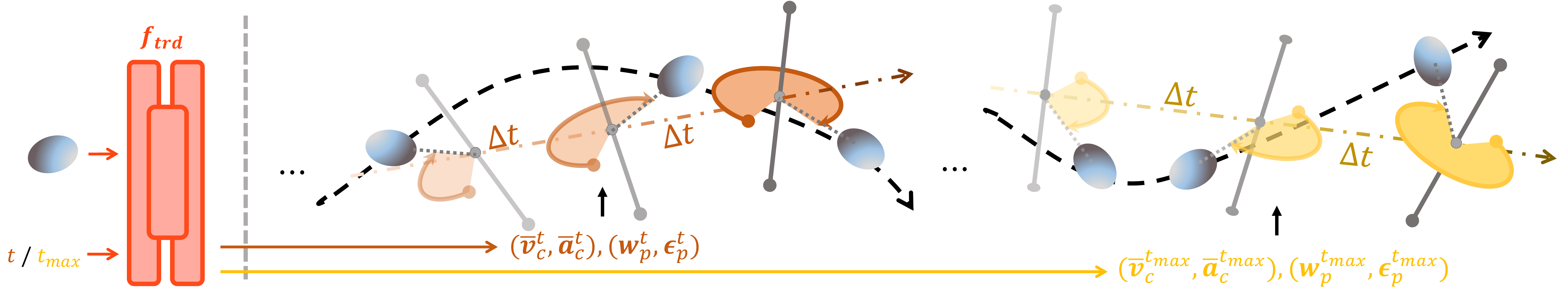}
    \vskip -0.1in
    \caption{The proposed translation rotation dynamics system for a specific rigid particle. The rigid particle will be driven by its learned physical parameters over time, forming a trajectory in 3D space.}
    \label{fig:trajectory}
    \vskip -0.2in
\end{figure*}

\textbf{Auxiliary Deformation Field}: To aid the learning of our translation and rotation dynamics system, we leverage an existing deformation field \citep{Yang2024}, but we are also amenable to other deformable Gaussian methods such as 4DGS \citep{Wu2024}, as demonstrated in our experiments in Section \ref{sec:main_exp_extrapolation}. In particular, the 3D position $\boldsymbol{x}_0$ of each canonical Gaussian kernel and the current timestamp $t$ are fed into an MLP-based deformation network, denoted as $f_{defo}$, directly predicting the corresponding position displacement $\delta \boldsymbol{x}$, and the change of quaternion $\delta \boldsymbol{r}$ and scaling $\delta \boldsymbol{s}$ from timestamp $0$ to $t$. All Gaussians $\boldsymbol{G}_t$ at time $t$ can be easily computed as follows, where the operations $\circ$ and $\odot$ follow \citep{Yang2024}. 
\begin{equation}\label{eq:Gt}
\setlength{\abovedisplayskip}{3pt}
\setlength{\belowdisplayskip}{3pt}
    \boldsymbol{x}_t =\boldsymbol{x}_0 + \delta \boldsymbol{x}, \; \boldsymbol{r}_t=\boldsymbol{r}_0 \circ \delta \boldsymbol{r}, \; \boldsymbol{s}_t=\boldsymbol{s}_0 \odot \delta \boldsymbol{s}, \; \sigma, \; \boldsymbol{c}
\end{equation}

All these deformed Gaussians will be projected and optimized by visual images at timestamp $t$ in a later stage, as clarified in Section \ref{sec:train}, where the deformation net $f_{defo}$ will be optimized from scratch. 
All details are provided in Appendix \ref{sec:vanilla_gaussian} and \ref{sec:vanilla_deformation}.

\subsection{Translation Rotation Dynamics System}\label{sec:transRot}

This module aims to learn physical parameters that govern the motion of 3D scenes. However, the dynamics of an entire space are extremely complex. Here, we simplify this problem and formulate it into just learning per rigid particle dynamics, where we treat each (canonical or deformed) Gaussian kernel as a rigid particle with size and orientation. According to the laws of classical mechanics, for a specific rigid particle $\boldsymbol{P}\in \mathcal{R}^3$, its motion in a 3D world coordinate system can be regarded as a rotational movement about a rotation center which has its own translation. To this end, we aim to learn the following two groups of physical parameters for each 3D rigid particle $\boldsymbol{P}$: 
\begin{itemize}[leftmargin=*] 
\setlength{\itemsep}{1pt}
\setlength{\parsep}{1pt}
\setlength{\parskip}{1pt}
    \item Group \#1 - Rotation Center Parameters including: 1) the center's position $\boldsymbol{P}_c\in \mathcal{R}^3$, 2) the center's velocity $\boldsymbol{v}_c\in \mathcal{R}^3$, and 3) acceleration $\boldsymbol{a}_c\in \mathcal{R}^3$ in world coordinate.
    \item Group \#2 - Rigid Particle Rotational Parameters including: 1) the rigid particle's rotation vector $\boldsymbol{w}_p \in \mathcal{R}^3$ with regard to its center $\boldsymbol{P}_c$, and 2) the rigid particle's angular acceleration $\boldsymbol{\epsilon}_p \in \mathcal{R}^3$. 
\end{itemize} 

As illustrated in Figure \ref{fig:trajectory}, our translation rotation dynamics system module, denoted as $f_{trd}$, takes a rigid particle $\boldsymbol{P}$ and a time $t$ as input, directly predicting the physical parameters of its rotation center and its own rotational information. Then, this rigid particle will be driven by its learned physical parameters, forming its motion dynamics, as illustrated by the trajectory in Figure \ref{fig:trajectory}. The composite velocity for the rigid particle $\boldsymbol{P}$ at time $t$ is derived as follows.
\begin{equation}
\label{eqn:vel_ambiguity}
\setlength{\abovedisplayskip}{3pt}
\setlength{\belowdisplayskip}{3pt}
    \boldsymbol{v}_{p}^t = \boldsymbol{w}_p^t \times (\boldsymbol{P} - \boldsymbol{P}_c^t) + \boldsymbol{v}_c^t = \boldsymbol{w}_p^t \times \boldsymbol{P} + (\boldsymbol{v}_c^t - \boldsymbol{w}_p^t \times \boldsymbol{P}_c^t)
\end{equation}

Nevertheless, the rightmost block of above equation shows that the estimation of the  center's velocity $\boldsymbol{v}_c^t$ and center's position $\boldsymbol{P}_c^t$ is compounded. Thus, instead of separately learning these two parameters, we turn to learn an equivalent (compounded) center velocity $\Bar{\boldsymbol{v}}_c^t$. Similarly, we instead learn an equivalent center acceleration $\Bar{\boldsymbol{a}}_c^t$:
\begin{equation}
\label{eqn:vel_equiv}
\setlength{\abovedisplayskip}{3pt}
\setlength{\belowdisplayskip}{3pt}
    \Bar{\boldsymbol{v}}_{c}^t = \boldsymbol{v}_c^t - \boldsymbol{w}_p^t \times \boldsymbol{P}_c^t, \quad \quad \quad \Bar{\boldsymbol{a}}_{c}^t = \boldsymbol{a}_c^t - \boldsymbol{\epsilon}_p^t \times \boldsymbol{P}_c^t.
\end{equation}

More details about the definition of equivalent parameters and the proof for equivalence are provided in Appendix \ref{app:equivalence}. This $f_{trd}$ module is implemented by simple MLPs as:
\begin{equation}\label{eq:f_trd}
\setlength{\abovedisplayskip}{3pt}
\setlength{\belowdisplayskip}{3pt}
 \{(\Bar{\boldsymbol{v}}_c^t, \Bar{\boldsymbol{a}}_c^t), (\boldsymbol{w}_p^t, \boldsymbol{\epsilon}_p^t)\} = f_{trd}(\boldsymbol{P}, t)
\end{equation}

Notably, for an input rigid particle $\boldsymbol{P}$, the elegance of this module $f_{trd}$ is that it only needs to learn this full translation rotation dynamics system at one specific time $t$, and that particle's future motion will be governed by the learned dynamics system according to the laws of mechanics. 

Due to the complex change of direction of rotation vector $\boldsymbol{w}_p$ driven by the angular acceleration $\boldsymbol{\epsilon}_p$ for a particle, we use Runge-Kutta 2$^{nd}$-order method to numerically calculate the evolving dynamics 
to derive the future parameters, as detailed in Algorithm \ref{alg:rk2}.

Instead of using 2$^{nd}$-order updating scheme, we can also extend to higher orders or reduce to lower orders with regard to future time. Intuitively, a higher order relationship
is expected to capture extremely complex dynamics such as a rolling ball suddenly breaking up into pieces due to unknown explosives inside, whereas a much lower order relationship tends to only capture static or constant speed scenes, thus being oversimplified. In this paper, we opt for the above 2$^{nd}$-order scheme to update dynamics parameters
for two primary reasons:
\begin{itemize}[leftmargin=*] 
\setlength{\itemsep}{1pt}
\setlength{\parsep}{1pt}
\setlength{\parskip}{1pt}
\item In many applications such as robot manipulation, the need for future prediction usually involves a relatively short interval, \eg{}, in milliseconds. In this case, a 2$^{nd}$-order update is usually sufficient to achieve decent approximations. Additionally, a simple sliding window based approach can be applied to continuously predict future frames given the newest visual observations from sensors. 
      
\item In our daily life, the majority of common physical movements such as rolling balls or moving cars can be generally described by a 2$^{nd}$-order relationship. In fact, both Newton's First and Second Law of Motion can be captured. Notably, since the whole 3D scene comprises a large number of rigid particles, each has up to 2$^{nd}$-order dynamics. Therefore, the compounded dynamics for the entire 3D scene can be rather complex, including various deformations and transformations in our daily lives.     
\end{itemize} 
Nevertheless, it is still interesting yet non-trivial to learn much higher-order relationships, and we leave it for future exploration. Implementation details are in Appendix \ref{sec:transRotSystem}.

\subsection{Training}\label{sec:train}
With our translation rotation dynamics module and the auxiliary deformation field, we now discuss how to train them together, such that physical parameters can be truly learned. 

\textbf{First}, for two timestamps $t'$ and $t$, where $t$ is usually sampled from the training set and $\Delta t = t - t'$ is predefined to be small enough, we can easily obtain Gaussians $G_{t'}$ from the deformation field $f_{defo}$ based on Equation \ref{eq:Gt}.

\textbf{Second}, having our translation rotation dynamics module $f_{trd}$ at hand, we naturally regard the transportation of all kernels from $t'$ to $t$ is governed by the corresponding physical parameters estimated by $f_{trd}$ at time $t'$ as defined in Equation \ref{eq:f_trd}. Then we use the Runge-Kutta 2$^{nd}$-order (RK2) method to numerically derive the future physical parameters. Thus we can easily compute the kernel's orientation $\boldsymbol{r}_t$ from $\boldsymbol{r}_{t'}$ and the kernel's position translation $\boldsymbol{x}_t$ from $\boldsymbol{x}_{t'}$. The kernel's translation consists of two parts: 1) the equivalent translation of its rotation center, and 2) the equivalent displacement caused by the kernel's rotation with regard to its center. In particular, the RK2 method is defined as:
\begin{algorithm}[H] 
\caption{ {\small Runge-Kutta 2$^{nd}$-order (RK2) method}
}
\label{alg:rk2}
\begin{algorithmic} 
\footnotesize
\State{\textbf{Input:}} 
(1) Gaussians at $t'$: $\bigl\{\cdot\cdot (\boldsymbol{x}_{t'}, \boldsymbol{r}_{t'}, \boldsymbol{s}_{t'}, \sigma, \boldsymbol{c}) \cdot\cdot \bigr\}$, and (2) $\Delta t$ 
    
\State{\textbf{Output:}} 
Gaussians at $t=(t' + \Delta t)$: $\bigl\{\cdot\cdot (\boldsymbol{x}_{t}, \boldsymbol{r}_{t}, \boldsymbol{s}_{t}, \sigma, \boldsymbol{c}) \cdot\cdot \bigr\}$

\State{\phantom{xx}$\bullet\   \{(\Bar{\boldsymbol{v}}_c^{t'}, \Bar{\boldsymbol{a}}_c^{t'}), (\boldsymbol{w}_p^{t'}, \boldsymbol{\epsilon}_p^{t'})\} \leftarrow f_{trd}(\boldsymbol{x}_{t'}, t')$, (ref to Eq \ref{eq:f_trd})}

\State{\phantom{xx}$\bullet\ \Bar{\boldsymbol{v}}_c^{mid} \leftarrow \Bar{\boldsymbol{v}}_c^{t'} + \frac{\Delta t}{2} \Bar{\boldsymbol{a}}_c^{t'}$}

\State{\phantom{xx}$\bullet\ \boldsymbol{w}_p^{mid} \leftarrow \boldsymbol{w}_p^{t'} + \frac{\Delta t}{2} \boldsymbol{\epsilon}_p^{t'}$}

\State{\phantom{xx}$\bullet\ \boldsymbol{x}_t \leftarrow \boldsymbol{x}_{t'} + \Delta t (\Bar{\boldsymbol{v}}_c^{mid} + \boldsymbol{w}_p^{mid}\times \boldsymbol{x}_{t'})$}

\State{\phantom{xx}$\bullet$ \text{Convert quaternion to rotation matrix:} $\mathbf{R}_{t'} \leftarrow \boldsymbol{r}_{t'}$}

\State{\phantom{xx}$\bullet$ Calculate the Cross-Product Matrix, denoted by $\boldsymbol{W}_p \in \mathcal{R}^{3\times3}$, for the normalized vector $\boldsymbol{w}_p^{mid}/\|\boldsymbol{w}_p^{mid}\|$. Details are in Appendix \ref{app:cross-product}.}

\State{\phantom{xx}$\bullet\  \Delta\mathbf{R} \leftarrow \mathbf{I} + (\sin{\Delta\theta})\boldsymbol{W}_p + (1-\cos{\Delta\theta})\boldsymbol{W}_p^2$, where $\Delta\theta = \Delta t \boldsymbol{w}_p^{mid}$, following Rodrigue's Formula \citep{rodrigues1840}}

\State{\phantom{xx}$\bullet\ \mathbf{R}_{t} \leftarrow (\Delta\mathbf{R})\mathbf{R}_{t'}$}

\State{\phantom{xx}$\bullet$ \text{Convert rotation matrix to quaternion:} $\boldsymbol{r}_{t} \leftarrow \mathbf{R}_{t} $ }

\State{\phantom{xx}$\bullet$ \text{Assign $\boldsymbol{s}_{t'}$ to $\boldsymbol{s}_t$: $\boldsymbol{s}_t \leftarrow \boldsymbol{s}_{t'}$}, as each Gaussian is rigid with the same scale by definition} 

\\ \textbf{Return:} All Gaussians at $t$: $\bigl\{\cdot\cdot (\boldsymbol{x}_{t}, \boldsymbol{r}_{t}, \boldsymbol{s}_{t}, \sigma, \boldsymbol{c}) \cdot\cdot \bigr\}$
\end{algorithmic}
\end{algorithm}

\textbf{Third}, we render all the above Gaussians kernels at time $t$ to 2D image space following 3DGS, comparing with the training images at time $t$. All parameters are trained with: 
\begin{equation}
    (\boldsymbol{G}_0, f_{defo}, f_{trd}) \longleftarrow (\ell_1 + \ell_{ssim})
\end{equation}

\section{Experiments}
\label{sec:exp}
\begin{table*}[t]\tabcolsep=0.2cm  
\centering
\caption{Quantitative results of all methods for future frame extrapolation on all four datasets.} \vspace{-0.2cm}
\resizebox{1\linewidth}{!}{
\begin{tabular}{r|ccc|ccc|ccc|ccc}
\toprule 
 & \multicolumn{3}{c|}{Dynamic Object} & \multicolumn{3}{c|}{Dynamic Indoor Scene} & \multicolumn{3}{c|}{NVIDIA Dynamic Scene} & \multicolumn{3}{c}{Dynamic Multipart} \\[+0.1em] 
 \cmidrule{2-13} 
 & PSNR$\uparrow$ & SSIM$\uparrow$ & \multicolumn{1}{c|}{LPIPS$\downarrow$} & PSNR$\uparrow$ & SSIM$\uparrow$ & LPIPS$\downarrow$ & PSNR$\uparrow$ & SSIM$\uparrow$ & \multicolumn{1}{c|}{LPIPS$\downarrow$} & PSNR$\uparrow$ & SSIM$\uparrow$ & LPIPS$\downarrow$ \\ 
 \midrule
T-NeRF\citep{Pumarola2021} & 13.818 & 0.739 & 0.324 & 22.242 & 0.700 & 0.363 & 21.120 & 0.707 & 0.358 & 10.064 & 0.576 & 0.537 \\
D-NeRF\citep{Pumarola2021} & 14.660 & 0.737 & 0.312 & 20.791 & 0.692 & 0.349 & 20.633 & 0.709 & 0.327 & 13.344 & 0.767 & 0.340 \\
TiNeuVox\citep{Fang2022} & 19.612 & 0.940 & 0.073 & 21.029 & 0.770 & 0.281 & 24.556 & 0.863 & 0.215 & 20.804 & 0.923 & 0.090  \\
T-NeRF$_{PINN}$ & 16.189 & 0.835 & 0.230 & 17.290 & 0.477 & 0.618 & 17.975 & 0.605 & 0.428 & - & - & -  \\
HexPlane$_{PINN}$ & 21.419 & 0.946 & 0.067 & 23.091 & 0.742 & 0.401 & 24.473 & 0.818 & 0.279 & - & - & -  \\
NVFi\citep{Li2023c}  & 27.594 & 0.972 & 0.036 & 29.745 & 0.876 & 0.204 & \underline{28.462} & 0.876 & 0.214 & 25.235 & 0.955 & 0.046  \\
DefGS\citep{Yang2024} & 19.849 & 0.949 & 0.045 & 21.380 & 0.819 & 0.188 & 24.240 & 0.895 & 0.140 & 20.664 & 0.930 & 0.067  \\ 
DefGS$_{nvfi}$ & 28.749 & \underline{0.984} & \underline{0.013} & \underline{31.096} & \underline{0.945} & \underline{0.077} & 27.529 & \underline{0.927} & \underline{0.102} & 28.455 & 0.979 & 0.017  \\ 
4DGS\citep{Wu2024} & 20.354 & 0.950 & 0.052 & 21.107 & 0.793 & 0.274 & 22.510 & 0.703 & 0.408 & 20.564 & 0.935 & 0.067 \\[+0.1em] 
\midrule 
\textbf{\nickname{}$_{4dgs}$ (Ours)} & \underline{30.327} & 0.983 & 0.019 & 30.607 & 0.896 & 0.209 & 22.554 & 0.721 & 0.390 & \underline{32.317} & \underline{0.985} & \underline{0.016}\\
\textbf{\nickname{} (Ours)} & \textbf{31.597} & \textbf{0.987} & \textbf{0.009} & \textbf{34.824} & \textbf{0.965} & \textbf{0.054} & \textbf{29.341} & \textbf{0.933} & \textbf{0.074} & \textbf{33.481} & \textbf{0.990} & \textbf{0.007}\\[+0.1em] 
\bottomrule
\end{tabular}
}
\label{tab:exp_extrapolation_only}
\vspace{-0.6cm}
\end{table*}

\paragraph{Datasets:} Our method is designed to learn meaningful physical information of 3D dynamic scenes, aiming at accurately predicting future motions, instead of just fitting observed video frames. In this regard, the closest work to us is the recent NVFi \citep{Li2023c}. Following NVFi, we primarily evaluate our method on its three dynamic datasets: \textbf{1) Dynamic Object dataset} consisting of 6 dynamic objects. Each object displays a unique motion pattern belonging to either rigid or deformable movement. \textbf{2) Dynamic Indoor Scene dataset} with 4 complex indoor scenes. Each scene has multiple objects undergoing different rigid body motions. \textbf{3) NVIDIA Dynamic Scene dataset} \citep{Yoon2020}. It consists of two chosen real-world dynamic 3D scenes. 

Upon a closer look at the above three datasets, we find that their dynamics captured are relatively simple. In our daily life, the majority of objects and scenes have multiple parts undergoing radically different motions over time, showing challenging physical patterns to learn. 
To further evaluate the effectiveness of our design, we collect a new synthetic dataset, named \textbf{4) Dynamic Multipart dataset}. 

Our new synthetic dataset comprises 4 objects. Each has 2 to 5 distinct motion patterns on different object parts. Following \citep{Li2023c}, for each object, we collect RGBs at 15 different viewing angles, where each viewing angle has 60 frames captured. We reserve the first 46 frames at randomly picked 12 viewing angles as the training split, \ie{}, 552 frames, while leaving the 46 frames at the remaining 3 viewing angles for testing \textit{interpolation} ability, \ie{}, 138 frames for novel view synthesis within the training time period, and keeping the last 14 frames at all 15 viewing angles for evaluating future frame \textit{extrapolation}, \ie{}, 210 frames.

\textbf{Baselines:} We select the following baselines: \textbf{1) NVFi} \citep{Li2023c}: This is the closest work to us, but differs from us in two folds. First, NVFi relies on PINN losses to learn physics priors, but we directly learn physical parameters. Second, NVFi adopts NeRF as a backbone, being short in 3D scene geometry and appearance modeling, but our method is amenable to and adopts the powerful 3DGS in nature. \textbf{2) T-NeRF} \citep{Pumarola2021}. \textbf{3) D-NeRF} \citep{Pumarola2021}. 
\textbf{4) TiNeuVox} \citep{Fang2022}. The latter four methods are based on NeRF and designed for novel view interpolation. Therefore they are expected to be rather weak for future frame extrapolation. For a fair and extensive comparison, we also include the following two baselines. \textbf{5) DefGS} \citep{Yang2024}, and \textbf{6) 4DGS} \citep{Wu2024}. Both very recent deformable 3D Gaussians methods are particularly strong to model dynamic 3D scenes for novel view synthesis using 3DGS as a backbone. \textbf{7) DefGS$_{nvfi}$}. We build this baseline by combining DefGS with the velocity field proposed by NVFi. This baseline has the powerful 3DGS as a backbone as well as the current state-of-the-art NVFi learning strategy. It is trained with exactly the same settings as our method. To demonstrate the flexibility of our framework, we also adopt 4DGS as our auxiliary deformation field, denoted as \nickname{}$_{4dgs}$.
 
\textbf{Metrics:} We report \textbf{PSNR}/ \textbf{SSIM}/ \textbf{LPIPS} scores for RGB synthesis in future frame extrapolation.

\subsection{Main Results of Future Frame Extrapolation}\label{sec:main_exp_extrapolation}

All methods are trained in a scene-specific fashion. 
As a common practice, the actual time length in all datasets is normalized to be one time unit, where the first $(0\sim 0.7)$ time unit is training time and the remaining is test time. In training and test, we set $\Delta t$ to be 2 divided by the training set frame rate. To extrapolate future frames, the time $t'$ for our auxiliary deformation field $\fdefo$ is set as $0.7$.
The future frames are progressively calculated by Algorithm \ref{alg:rk2} with the same time interval $\Delta t$.

Our primary goal is to extrapolate meaningful future frames as a continuum of the last training observations. In our evaluation, we follow the second step
in Section \ref{sec:train} to extrapolate future frames from the last timestamp of training frames. We also compare novel view (past frame) interpolation with baselines, but this is less important. Details of interpolation and the results are provided in Appendix \ref{app:full_main_table}.

\textbf{Results \& Analysis:} Table \ref{tab:exp_extrapolation_only} compare future frame extrapolation of all methods on four datasets. We can see that: 
\begin{itemize}[leftmargin=*] 
\setlength{\itemsep}{1pt}
\setlength{\parsep}{1pt}
\setlength{\parskip}{1pt}
\item Compared with NeRF and 3DGS based dynamic scene modeling methods such as T-NeRF/ D-NeRF/ TiNeuVox/ DefGS/ 4DGS, both versions of our method achieve about 10 points higher on PSNR for future frame extrapolation. This means that, without explicitly learning physical information like us, these methods completely fail to predict the future, highlighting the core value of our method.
\item Compared with the closest and also strongest baselines NVFi/ DefGS$_{nvfi}$, our method is still constantly better than them on all datasets for future frame extrapolation. Notably, on the Dynamic Indoor Scene dataset and our newly collected Dynamic Multipart dataset, there are much more complex motion dynamics such as different objects or parts moving in distinct directions, but our best results are constantly about 3 points higher on PSNR than them. Fundamentally, this is because both NVFi and DefGS$_{nvfi}$ rely on PINN losses as soft constraints to incorporate physics priors, whereas we directly integrate hard physics by learning translation rotation parameters, thus being more effective in learning dynamics. 
\item Lastly, our framework is indeed amenable to existing deformation fields such as DefGS and 4DGS, and both versions achieve superior results for future frame extrapolation on most datasets. 
\end{itemize} 
Figure \ref{fig:qual_res} shows qualitative results. More results are in Appendix \ref{sec:app_res_dynObj} / \ref{sec:app_res_indoorScene} /  \ref{sec:app_res_nvidia} / \ref{sec:app_res_multipart}. We also report the training/test time, memory cost, \etc{}, in Appendix \ref{sec:app_resources}.

\subsection{Analysis of Dynamics Parameters}

Our core translation rotation dynamics system module is designed to learn per rigid particle's physical parameters. Ideally, for those particles undergoing the same motion pattern such as all surface points of a single rigid part, they should have the same or similar physical parameters. Given this, multiple dynamic objects or parts with distinct motions can be automatically segmented based on the similarity of learned physical parameters. By comparison, the prior work NVFi \citep{Li2023c} can hardly achieve this autonomous dynamic segmentation by its own design, unless an additional motion grouping method is applied. 
To further evaluate this nice property of our method, we conduct the following steps to analyze the learned dynamics parameters. 

First, after training our method on the Dynamic Object dataset, Dynamic Multipart dataset, and Dynamic Indoor Scene dataset, for each dynamic scene, we have a set of well-trained canonical 3D Gaussians, an auxiliary deformation field, and our translation rotation dynamics parameters. 

Then, we use the auxiliary deformation field $\fdefo$ to deform the canonical 3D Gaussians $\boldsymbol{G}_0$ to time $t=0.7$, which is the maximum time the deformation field can query in our training. At this timestamp, the motions of different objects and parts normally achieve a steady state. 

\begin{figure}[ht]
\centering
  \includegraphics[width=0.99\linewidth]{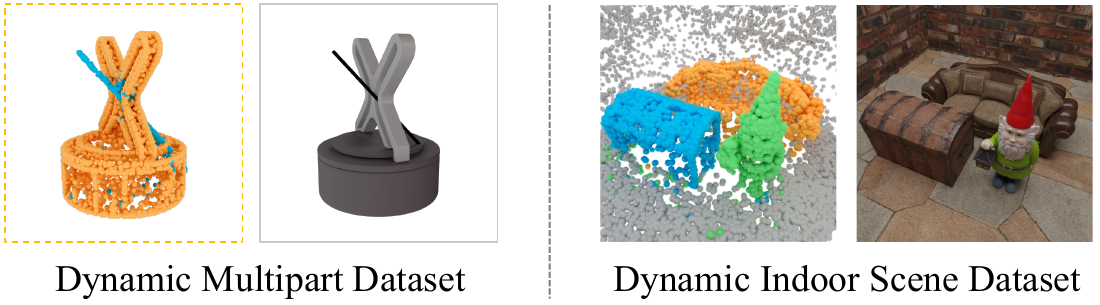}
  \vspace{-0.3cm}
\caption{Clustering of translation rotation physics parameters. 
}
\label{fig:segmentation}
\vspace{-0.3cm}
\end{figure}

Lastly, we query all the learned physical parameters at this time, \ie{}, $\{(\Bar{\boldsymbol{v}}_c^{t}, \Bar{\boldsymbol{a}}_c^{t}), (\boldsymbol{w}_p^{t}, \epsilon_p^{t})\}$. We choose $(\|\Bar{\boldsymbol{v}}_c^{t}\|, \Bar{\boldsymbol{v}}_c^{t}, \|\boldsymbol{w}_p^{t}\|, \boldsymbol{w}_p^{t}/\|\boldsymbol{w}_p^{t}\|)$ as the features to cluster Gaussians via a simple K-means algorithm. As shown in Figure \ref{fig:segmentation}, 
all Gaussians can be grouped into physically meaningful objects or parts according to their actual motion patterns. More results are in Appendix \ref{sec:app_res_seg_point}\ /\ \ref{sec:app_res_seg_mask}. 

We further quantitatively evaluate our motion grouping results on Dynamic Indoor Scene Dataset. In particular, we follow Gaussian Grouping \citep{gaussian_grouping} to render 2D object segmentation masks for all 30 views over 60 timestamps on all 4 scenes, \ie, 7200 images in total. We compare with \textbf{D-NeRF}, \textbf{NVFi}, \textbf{DefGS} and \textbf{DefGS$_{nvfi}$}. We follow NVFi to obtain segmentation results of D-NeRF and NVFi. For the 3DGS-based baselines, we also adopt OGC \citep{Song2022} to segment Gaussians based on scene flows induced from their learned deformation fields. All implementation details are in Appendix \ref{app:ogc_detail}. We also include a strong 2D object segmentation method, Mask2Former \citep{Cheng2022} pre-trained by human annotations on COCO \citep{Lin2014} as a fully supervised baseline.

As shown in Table \ref{tab:exp_decomposition}, our method achieves almost perfect object segmentation results on all metrics, significantly outperforming all baselines. This shows that our learned physical parameters correctly model object physical motion patterns and can be easily used to identify objects according to their motions, without needing any human annotations.

\setlength{\abovecaptionskip}{0 pt}
\begin{table}[t] \tabcolsep=0.2cm 
\caption{Quantitative results of motion segmentation results on Dynamic Indoor Scene dataset.}
\label{tab:exp_decomposition}
\footnotesize
\resizebox{1.\linewidth}{!}{
\begin{tabular}{rcccccc}
\hline
            & AP$\uparrow$    & PQ$\uparrow$    & F1$\uparrow$    & Pre$\uparrow$    & Rec$\uparrow$    & mIoU$\uparrow$  \\ \hline
M2F\citep{Cheng2022}   & 65.37 & 73.14 & 78.29 & \underline{94.83} & 68.88 & 64.42 \\
D-NeRF\citep{Pumarola2021} & 57.26 & 46.15 & 59.02 & 56.55 & 62.94 & 46.58 \\
NVFi\citep{Li2023c} & \underline{91.21} & \underline{78.74} & \underline{93.75} & 93.76 & \underline{93.74} & \underline{67.64} \\
DefGS\citep{Yang2024} & 51.73 & 57.60 & 66.43 & 63.21 & 70.07  & 54.46 \\
DefGS$_{nvfi}$ & 55.26 & 62.75 & 69.83 & 69.39 & 72.91 & 56.82  \\ \hline
\textbf{\nickname{} (Ours)} & \textbf{95.82} & \textbf{93.28} & \textbf{97.90} & \textbf{96.21} & \textbf{99.86} & \textbf{79.55} \\\hline
\end{tabular}
}
\vspace{-0.7cm}
\end{table}

\subsection{Continual Learning}

We include an additional continual learning experiment to show that our framework can easily adapt to new observations to learn rapidly changing dynamics. We test on \textbf{ParticleNeRF dataset} \cite{abou2022particlenerf}. It has 6 dynamic objects. Each object undergoes various rigid motions from robot manipulation to harmonic oscillation or deformable motions such as cloth. Details of the dataset are in Appendix \ref{app:incremental}.  

Particularly, we first feed frames of $(0\sim0.15)$ time unit to train the network, and evaluate  
future frame extrapolation on $(0.15\sim0.30)$. Next, we include $(0.15\sim0.30)$ to continue training, and then evaluate  
future frame extrapolation on $(0.30\sim0.45)$. We keep adding a time unit of 0.15 till training on $(0\sim0.75)$, and extrapolate on $(0.75\sim0.9)$. 

We compare our extrapolation performance with DefGS and DefGS$_{nvfi}$. Table \ref{tab:exp_incremental_} shows quantitative results. 
It can be seen that DefGS and DefGS$_{nvfi}$ struggle in making right future predictions based on novel observations, while our model can stably adapt to new observations and achieve excellent future frame predictions. This means that even though the 3D scene dynamics are radically and rapidly changing over time, our model can continue capturing those dynamics given new observations. We believe this would be a key enabler for robot perception and manipulation in highly dynamic environments. Qualitative extrapolation and past time interpolation results are in Appendix \ref{app:incremental}.
\begin{table}[ht]\tabcolsep=0.25cm  \vspace{-0.1cm}
\centering
\caption{Quantitative results (PSNR) of continual learning.}
\resizebox{1\linewidth}{!}{
\begin{tabular}{r|ccccc|c}
\toprule
Observed Time & 0.15 & 0.30 & 0.45 & 0.60 & 0.75 & \multirow{2}*{Average} \\ 
Extrap Till & 0.30 & 0.45 & 0.60 & 0.75 & 0.90 &  \\ 
 \midrule
DefGS\citep{Yang2024} & 19.856 & 18.652 & 20.141 & 20.813 & 19.131 & 19.718\\
DefGS$_{nvfi}$ & 26.105 & 25.987 & 27.122 & 26.051 & 25.640 & 26.181\\
\textbf{\nickname{} (Ours)} & \textbf{26.731} & \textbf{27.623} & \textbf{27.168} & \textbf{28.422} & \textbf{27.777} & \textbf{27.544}\\
\bottomrule
\end{tabular}
}
\label{tab:exp_incremental_}
\vspace{-0.5cm}
\end{table}

\begin{figure*}[t]
    \centering
    \includegraphics[width=1\linewidth]{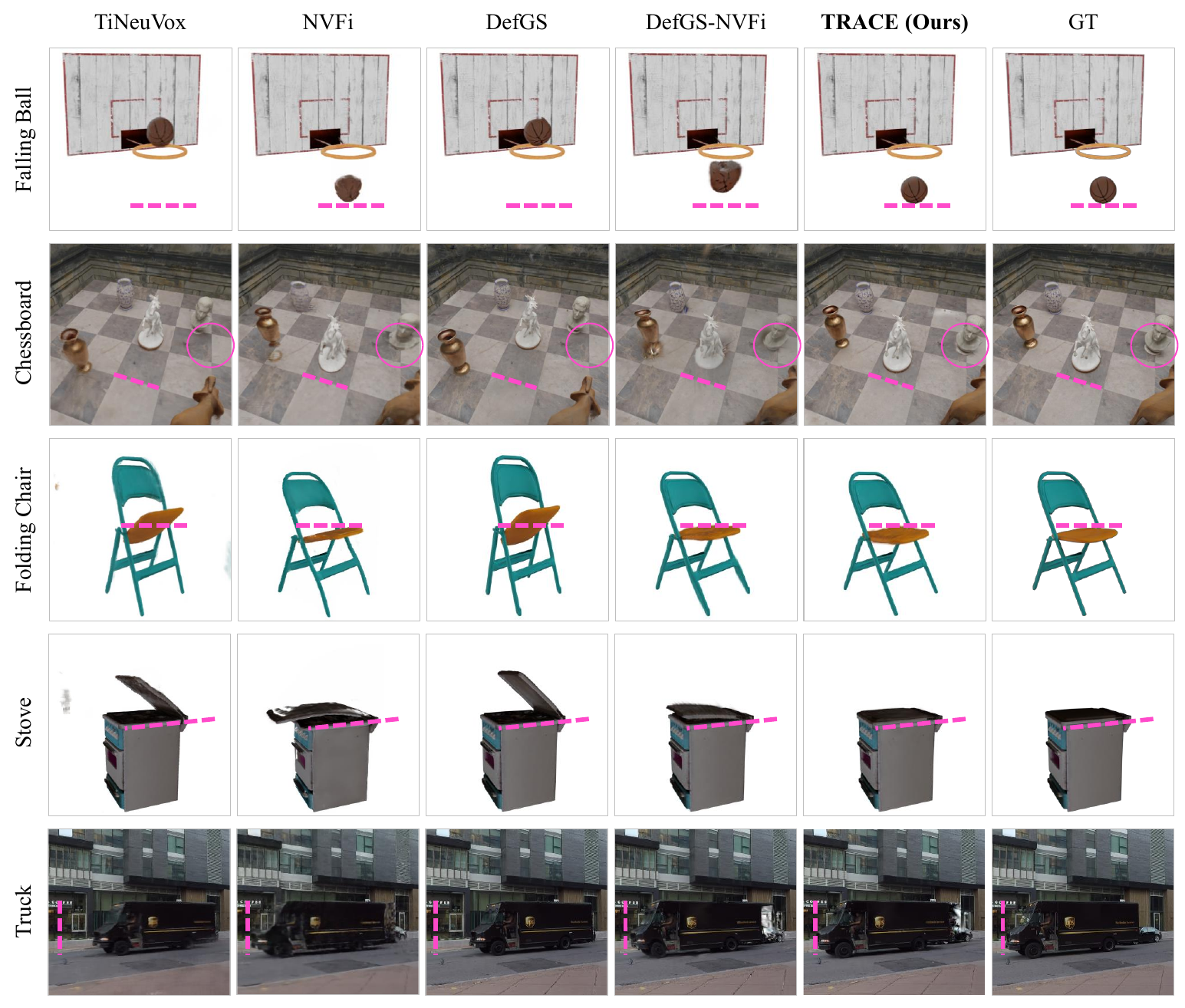}
    \caption{Qualitative results of RGB view synthesis for future frame extrapolation on four datasets.}
    \label{fig:qual_res} \vspace{-0.4cm}
\end{figure*}

\subsection{Ablation Study}

To verify different choices of our method, we conduct the following five groups of ablation experiments.   

\textbf{(1) Different choices of time difference $\Dt$ in training stage}: Given the time interval between two consecutive frames in the training set as $\delta t$, we compare three choices of the time difference $\Dt$ in the training stage: $\{\delta t, 2\delta t, 3\delta t\}$. We choose $\Dt = 2\delta t$ in our main experiments.

\textbf{(2) Different choices of the order of physical dynamics}: Our translation rotation dynamic systems choose to learn 2$^{nd}$-order dynamics in our main experiments. 
We compare the choices for the 1$^{st}$ order (no acceleration) and the 3$^{rd}$ order (acceleration of acceleration). 

\textbf{(3) Removing the auxiliary deformation field $f_{defo}$}: We feed Gaussian s at time $t'=0$ directly into $f_{trd}$, and use the output physical parameters to progressively move s to a target future time $t$ with $\Dt = 2\delta t$.

\textbf{(4) \textcolor{black}{Removing the physics derivation}}: Instead of using the physics parameters queried at time $t'$ to derive the future motion states at time $t > t'$, we query the physics parameters from $f_{\mathrm{trd}}(\boldsymbol{P}, t)$ for future timestamps.

\textbf{(5) Removing the equivalence parametrization:} Instead of learning the equivalent velocity and acceleration for the centers as Equation \ref{eqn:vel_equiv}, we directly learn the original parameters, implemented by a same MLP only with a modified output layer: $\{(\boldsymbol{P}_c^{t}, \boldsymbol{v}_c^{t}, \boldsymbol{a}_c^{t}), (\boldsymbol{w}_p^{t}, \epsilon_p^{t})\} = f_{\mathrm{trd'}}(\boldsymbol{P}, t')$. 

\setlength{\abovecaptionskip}{0 pt}
\begin{table}[ht] \tabcolsep=0.035cm  \vspace{-0.2cm}
\caption{Ablation studies on three datasets.}
\label{tab:exp_ablation}
\footnotesize
\resizebox{1\linewidth}{!}{
\begin{tabular}{lcccc|ccc|ccc|ccc}
\toprule
 &  &  &  &  & \multicolumn{3}{c|}{Dynamic Multipart} & \multicolumn{3}{c|}{Dynamic Object} & \multicolumn{3}{c}{Dynamic Indoor Scene} \\ \cmidrule{6-14} 
  & order & $f_{defo}$ & physcis & equiv & PSNR$\uparrow$ & SSIM$\uparrow$ & LPIPS$\downarrow$ & PSNR$\uparrow$ & SSIM$\uparrow$ & LPIPS$\downarrow$ & PSNR$\uparrow$ & SSIM$\uparrow$ & LPIPS$\downarrow$ \\
\midrule
(1) $\delta t$ & 2 & \Y & \Y & \Y & 32.852 & 0.989 & 0.010 & \textbf{31.597} & 0.987 & 0.009 & 33.831 & 0.958 & 0.078 \\ 
(1) $2\delta t$ & 2 & \Y & \Y & \Y & \textbf{33.481} & \textbf{0.990} & \textbf{0.007} & 31.511 & \textbf{0.988} & \textbf{0.006} & \textbf{34.824} & \textbf{0.965} & \textbf{0.054} \\
(1) $3\delta t$ & 2 & \Y & \Y & \Y & 33.003 & 0.989 & 0.008 & 29.787 & 0.983 & 0.011 & 34.778 & \textbf{0.965} & \textbf{0.054} \\
\midrule
(2) $2\delta t$ & 1 & \Y & \Y & \Y & 33.125 & 0.989 & 0.010 & 28.522 & 0.984 & 0.012 & 34.576 & 0.963 & 0.076 \\
(2) $2\delta t$ & 3 & \Y & \Y & \Y & 33.312 & 0.989 & 0.008 & 31.044 & 0.987 & 0.007 & 34.086 & 0.961 & 0.056\\
\midrule
(3) $2\delta t$ & 2 & \N & \Y & \Y & 19.206 & 0.907 & 0.120 & - & - & - & - & - & - \\
(4) $2\delta t$ & 2 & \Y & \N & \Y & 29.602 & 0.983 & 0.012 & - & - & - & - & - & - \\
(5) $2\delta t$ & 2 & \Y & \Y & \N & 29.986 & 0.985 & 0.012 & - & - & - & - & - & - \\ 
\bottomrule
\end{tabular}
}
\vspace{-0.3cm}
\end{table} 

\textbf{Results \& Analysis}: Table \ref{tab:exp_ablation} shows all ablation results for future frame extrapolation. To demonstrate the robustness of our hyperparameters selected, the first two groups of ablations are extensively conducted on three datasets. The remaining ablations are primarily conducted on our new Dynamic Multipart dataset. It can be seen that: 1) The greatest impact is caused by the removal of the deformation field $f_{defo}$. While this deformation field itself is unable to learn physics, it significantly aids our core translation rotation dynamics system module to learn physical parameters given motion information. 2) The choice of time difference $\Delta t$ is also important. Once it is as small as the interval between two consecutive frames, the performance drops, because the motion in short intervals is too subtle to be distinguished. For example, a rotation may be learned as a translation. However, if $\Delta t$ is too large, the appearance fitting could be sacrificed, so the performance is slightly weaker. 3) If physical parameters are learned but not derived, the lack of physics consistency will influence motion learning. 
Ablation results for interpolation are in Appendix \ref{sec:app_res_ablation}.

\vspace{-0.1cm}
\section{Conclusion}
\label{sec:conclusion}
\vspace{-0.1cm}
In this paper, we have demonstrated that complex motion dynamics can be explicitly learned just from multi-view RGB videos without needing additional human labels such as object types and masks. This is achieved by a new generic framework that simultaneously models 3D scene geometry, appearance and physics by extending the appealing 3D Gaussian Splatting technique. In contrast to existing works which usually rely on PINN losses as soft constraints to learn physics priors, we instead directly learn a complete set of physical parameters to govern the motion pattern of each 3D rigid particle in space via our core translation rotation dynamics system module. Extensive experiments on three public dynamic datasets and a newly created dynamic multipart dataset have shown the extraordinary performance of our method in the challenging task of future frame extrapolation over all baselines. In addition, the learned physical parameters can be directly used to segment objects or parts according to the similarity of parameters.

\noindent\textbf{Acknowledgment:} 
This work was supported in part by Research Grants Council of Hong Kong under Grants 25207822 \& 15225522 \& 15219125, in part by Otto Poon Charitable Foundation Smart Cities Research Institute (8-CDCQ), in part by Research Centre for Unmanned Autonomous Systems (1-CE3D), The Hong Kong Polytechnic University. 

{
\small
\bibliographystyle{ieeenat_fullname}
\bibliography{references}
}

\clearpage

\setcounter{page}{1}
\maketitlesupplementary

The appendix includes:
\begin{itemize}
    \item Rendering equation and preliminary for vanilla 3DGS.
    \item Implementation details of Auxiliary Deformation Field.
    \item Details of  Equivalent Velocity and Acceleration.
    \item Details for the definition of Cross-Product Matrix.
    \item Implementation details of translation and rotation system.
    \item Training and evaluation resources for the model.
    \item Full quantitative results for both interpolation and extrapolation tasks.
    \item Quantitative and qualitative results for continual learning on Panoptic Sports dataset.
    \item Implementation details for the dynamic segmentation.
    \item Segmentation on Real-world Scene.
    \item Additional results for continual learning and the details for Particle NeRF dataset.
    \item Additional details of datasets.
    \item Analysis on performance change along prediction time span.
    \item Analysis on performance against motion complexity.
    \item Additional quantitative results for ablation study in the main context.
    \item Additional per-scene quantitative \& qualitative results for both interpolation and extrapolation tasks.
    \item Additional qualitative results for motion segmentation.
    \item Additional qualitative results for extrapolation beyond dataset time span.
\end{itemize}

\subsection{Preliminary for vanilla 3DGS}
\label{sec:vanilla_gaussian}

3D Gaussian Splatting \citep{Kerbl2023} represents a 3D scene by a set of colored 3D Gaussian kernels. Specifically, each Gaussian kernel is parameterized by a 3D position $\boldsymbol{P} \in \R^3$, an orientation represented by a quaternion $\boldsymbol{r}$, and a scaling $\boldsymbol{s}$. By transforming the orientation $\boldsymbol{r}$ and scaling $\boldsymbol{s}$ into the rotation matrix $\mathbf{R}$ and scaling matrix $\mathbf{S}$, a 3D covariance matrix $\Sigma$ can be composed as $\Sigma = \mathbf{R} \mathbf{S} \mathbf{S}^T \mathbf{R}^T$. Then the Gaussian kernel can be evaluated at any location $\boldsymbol{x} \in \R^3$ in the 3D space:
\begin{equation}
    G(\boldsymbol{x}) = e^{-\frac{1}{2}{(\boldsymbol{x}-\boldsymbol{P})}^T \Sigma^{-1} (\boldsymbol{x}-\boldsymbol{P})}.
\end{equation}
Besides, each Gaussian kernel has an opacity $\sigma$ indicating its influence in rendering, and a color $\boldsymbol{c}$ computed from spherical harmonics (SH) for view-dependent appearance. 

The rendering of Gaussian kernels on the image consists of two steps. Firstly, the Gaussian kernels are projected onto the image plane, following the differentiable rasterization pipeline proposed in \citep{Zwicker2001}. The 3D position $\boldsymbol{P}$ and covariance matrix $\Sigma$ of each Gaussian kernel are projected into 2D position $\boldsymbol{P}'=JW\boldsymbol{P}$ and covariance matrix $\Sigma'=JW\Sigma W^T J^T$ respectively, where $J$ denotes the Jacobian of the approximated projective transformation and $W$ denotes the transformation from the world to camera coordinates. Secondly, the color of a pixel $\boldsymbol{\mu}$ on the image can be rendered by $\alpha$-blending as follows:
\begin{equation}
    \boldsymbol{C}(\boldsymbol{\mu}) = \sum_{i} T_i \alpha_i \boldsymbol{c}_i, \quad T_i = \prod_{j=1}^{i-1} (1-\alpha_j),
\end{equation}
where $\alpha_i$ is obtained by evaluating the projection of the Gaussian kernel $G_i$ on the pixel $\boldsymbol{\mu}$, \ie, $\alpha_i = \sigma_i e^{-\frac{1}{2}{(\boldsymbol{\mu}-\boldsymbol{P'})}^T \Sigma'^{-1} (\boldsymbol{\mu}-\boldsymbol{P'})}$. By adjusting the parameters of Gaussian kernels mentioned above and adaptively controlling the Gaussian density, a high-fidelity representation of a 3D scene can be obtained from multi-view images. We refer readers to \citep{Kerbl2023} for more details.

\subsection{Implementation Details of Auxiliary Deformation Field}
\label{sec:vanilla_deformation}
We leverage an existing deformation field introduced in \citep{Yang2024} as our auxiliary deformation field. In particular, the 3D position $\x_0$ of each canonical Gaussian kernel and the current timestamp $t$ are fed into an MLP-based deformation network, denoted as $\fdefo$. The implementation of this MLP is directly adapted from \citep{Yang2024}, \ie{}, an MLP with 8 layers in total and 256 hidden sizes for each layer, plus a ResNet layer at layer 4. At the input layer, an 8-degree positional embedding is applied to the 3D position $\x_0$ and a 5-degree positional embedding onto time $t$. 

Mathematically, the deformation field $\fdefo(\x, t): \R^4 \rightarrow \R^{10}$ is defined as 

\begin{equation}
   (\delta\x, \delta\q, \delta\s) = \fdefo(\x_0, t),
\end{equation}

where $\delta\x$ represents the translation of the center of Gaussian kernel, $\delta\q$ represents the rotation for the pose of Gaussian kernel in quaternion representation, and $\delta\s$ is the difference of Gaussian sizes. Note a Gaussian size vector is parametrized by $\fs=\log(\s)$, so the difference can be defined as $\delta\s=\exp(\delta\fs)$. 

After applying the deformation field onto Gaussian kernels, we can deform the Gaussian kernels from canonical time to time $t$ as:

\begin{align}
   \x_t & = \x_0 + \delta\x \\
   \q_t & = \delta\q \circ \q_0 \\
   \s_t & = \mathrm{exp}(\fs_0 + \delta\fs)=\s_0\odot\delta\s,
\end{align}

where $\circ$ is quaternion multiplication and $\odot$ is element-wise multiplication.

\subsection{Details of Equivalent Velocity and Acceleration}\label{app:equivalence}

Given a rigid particle $\boldsymbol{P}$, we define the following physical parameters:
\begin{itemize}[leftmargin=*] 
\setlength{\itemsep}{1pt}
\setlength{\parsep}{1pt}
\setlength{\parskip}{1pt}
    \item Group \#1 - Rotation Center Parameters including: 1) the center's position $\boldsymbol{P}_c\in \mathcal{R}^3$, 2) the center's velocity $\boldsymbol{v}_c\in \mathcal{R}^3$, and 3) acceleration $\boldsymbol{a}_c\in \mathcal{R}^3$ in world coordinate.
    \item Group \#2 - Rigid Particle Rotational Parameters including: 1) the rigid particle's rotation vector $\boldsymbol{w}_p \in \mathcal{R}^3$ with regard to its center $\boldsymbol{P}_c$, and 2) the rigid particle's angular acceleration $\boldsymbol{\epsilon}_p \in \mathcal{R}^3$. 
\end{itemize} 

Then we can derive the composite velocity for this particle as:
\begin{equation}
\setlength{\abovedisplayskip}{3pt}
\setlength{\belowdisplayskip}{3pt}
    \boldsymbol{v}_{p}^t = \boldsymbol{w}_p^t \times (\boldsymbol{P} - \boldsymbol{P}_c^t) + \boldsymbol{v}_c^t = \boldsymbol{w}_p^t \times \boldsymbol{P} + (\boldsymbol{v}_c^t - \boldsymbol{w}_p^t \times \boldsymbol{P}_c^t).
\end{equation}

Now we define the equivalent velocity and acceleration as:
\begin{equation}
\setlength{\abovedisplayskip}{3pt}
\setlength{\belowdisplayskip}{3pt}
    \Bar{\boldsymbol{v}}_{c}^t = \boldsymbol{v}_c^t - \boldsymbol{w}_p^t \times \boldsymbol{P}_c^t, \quad \quad \quad \Bar{\boldsymbol{a}}_{c}^t = \boldsymbol{a}_c^t - \boldsymbol{\epsilon}_p^t \times \boldsymbol{P}_c^t.
\end{equation}

By definition, the $1^{st}$-order equivalence is naturally obeyed by the definition. To be specific, it can be shown as:
\begin{equation}
\setlength{\abovedisplayskip}{3pt}
\setlength{\belowdisplayskip}{3pt}
    \Bar{\boldsymbol{v}_{p}^t} = \boldsymbol{w}_p^t \times \boldsymbol{P} + \Bar{\boldsymbol{v}}_{c}^t = \boldsymbol{v}_{p}^t
\end{equation}

Now let's prove the $2^{st}$-order equivalence:
\begin{align}
    \Bar{\boldsymbol{v}}_{p}^{t+dt} & = (\boldsymbol{w}_p^t + dt\,\boldsymbol{\epsilon}_p^t) \times \boldsymbol{P} + \Bar{\boldsymbol{v}}_{c}^t + dt\,\Bar{\boldsymbol{a}}_{c}^t \\
    & = (\boldsymbol{w}_p^t + dt\,\boldsymbol{\epsilon}_p^t) \times \boldsymbol{P} + \boldsymbol{v}_c^t - \boldsymbol{w}_p^t \times \boldsymbol{P}_c^t \\ 
    & \quad\quad + dt\,(\boldsymbol{a}_c^t - \boldsymbol{\epsilon}_p^t \times \boldsymbol{P}_c^t) \\
    & = (\boldsymbol{w}_p^t + dt\,\boldsymbol{\epsilon}_p^t) \times \boldsymbol{P} - (\boldsymbol{w}_p^t + dt\,\boldsymbol{\epsilon}_p^t) \times \boldsymbol{P}_c^t \\
    & \quad\quad +  \boldsymbol{v}_c^t + dt\,\boldsymbol{a}_c^t \\
    & =  (\boldsymbol{w}_p^t + dt\,\boldsymbol{\epsilon}_p^t) \times (\boldsymbol{P} - \boldsymbol{P}_c^t) + (\boldsymbol{v}_c^t + dt\,\boldsymbol{a}_c^t) \\ 
    & = \boldsymbol{v}_{p}^{t+dt}
\end{align}

\subsection{Details of Cross-Product Matrix}\label{app:cross-product}
Here we first introduce the definition of Cross-Product Matrix. Given two vectors $\boldsymbol{k}=[k_1,k_2,k_3]^T\in\mathcal{R}^3$ and $\boldsymbol{v}=[v_1,v_2,v_3]\in\mathcal{R}^3$, one can calculate the cross product of these two vectors $\boldsymbol{k}\times\boldsymbol{v}$. Now we define the Cross-Product Matrix of vector $\boldsymbol{k}$ as $\boldsymbol{K}\in\mathcal{R}^{3\times3}$, such that
\begin{align}
     & \quad\quad\boldsymbol{K}\boldsymbol{v} = \boldsymbol{k}\times\boldsymbol{v} \\
     & =
     \begin{bmatrix}
        -k_3v_2 + k_2v_3\\
        k_3v_1 - k_1v_3\\
        -k_2v_1+k_1v_2
     \end{bmatrix}  =
     \begin{bmatrix}
         0 & -k_3 & k_2\\
         k_3 & 0 & k_1\\
         -k_2 & k_1 & 0
     \end{bmatrix}\boldsymbol{v}.
\end{align}
Thereby we can get the closed-form for the Cross-Product Matrix of vector $\boldsymbol{k}=[k_1,k_2,k_3]^T$ as 
\begin{equation}\label{eq:cross-product-matrix}
    \boldsymbol{K} = \begin{bmatrix}
         0 & -k_3 & k_2\\
         k_3 & 0 & k_1\\
         -k_2 & k_1 & 0
     \end{bmatrix}
\end{equation}

Now we want to derive the cross-product matrix for vector $\boldsymbol{w}_p^{mid}/\|\boldsymbol{w}_p^{mid}\|$. We just need to plug the vector components into the above Equation \ref{eq:cross-product-matrix}.

\subsection{Implementation details of translation and rotation system}\label{sec:transRotSystem}
As discussed in the main context, we learn the following two groups of physical parameters for each 3D particle $\boldsymbol{P}$ as function $\ftrd(\x): \R^3 \rightarrow \R^{13}$: 
\begin{equation}
\setlength{\abovedisplayskip}{3pt}
\setlength{\belowdisplayskip}{3pt}
 \{(\Bar{\boldsymbol{v}}_c, \Bar{\boldsymbol{a}}_c), (\boldsymbol{w}_p, \epsilon_p)\} = \ftrd(\boldsymbol{P}),
\end{equation}
where angular velocity is defined as $\omega = \|\boldsymbol{w}_p\|_2$ around the rotation axes direction $\boldsymbol{\hat{k}}=\boldsymbol{w}_p/\omega$ following the right-hand rule.

This module is implemented by a simple $8\times 256$ MLPs with 8 layers in total and 256 hidden sizes for each layer. In addition, an 8-degree positional embedding is applied onto the input 3D position $\x$, and $\mathrm{ReLU}$ is chosen as the activation function.

\subsection{Training and evaluation resources for the model}
\label{sec:app_resources}
As the complexity of different scenes varies, the total number of Gaussians learned for each scene varies from 40k to 1.6M. In general, our training time is 1.05 times longer than DefGS (or 4DGS if built on it). For example, on the \textit{bat} of Dynamic Object Dataset, DefGS/4DGS need 25 minutes, while we need 27 minutes, with a slight training cost addition. Since our additional module is a tiny MLPs, we only need 367.4kB larger storage. Our rendering speed is 0.85 times slower than DefGS (or 0.8 times slower than 4DGS if built on it). For example, on the \textit{bat} of Dynamic Object Dataset, they achieve 40 fps and ours 32 fps. We train all our models on a single NVIDIA 3090 24G GPU.

\subsection{Full Quantitative Results for both interpolation and extrapolation tasks}
\label{app:full_main_table}

We show the complete quantitative results for both novel-view \textit{interpolation} and future \textit{extrapolation} tasks in Table \ref{tab:exp_extrapolation} and Table \ref{tab:exp_extrapolation_real_multipart}. 

\begin{table*}[t]\tabcolsep=0.1cm 
\centering
\caption{Quantitative results of all methods for both future frame extrapolation and novel view interpolation on Dynamic Object Dataset and Dynamic Indoor Scene Dataset.}
\resizebox{1\linewidth}{!}{
\begin{tabular}{r|ccc|ccc|ccc|ccc}
\toprule
\multirow{3}{*}{} & \multicolumn{6}{c|}{Dynamic Object Dataset} & \multicolumn{6}{c}{Dynamic Indoor Scene Dataset} \\ \cmidrule{2-13} 
 & \multicolumn{3}{c|}{Interpolation} & \multicolumn{3}{c|}{Extrapolation} & \multicolumn{3}{c|}{Interpolation} & \multicolumn{3}{c}{Extrapolation} \\ \cmidrule{2-13} 
 & PSNR$\uparrow$ & SSIM$\uparrow$ & \multicolumn{1}{c|}{LPIPS$\downarrow$} & PSNR$\uparrow$ & SSIM$\uparrow$ & LPIPS$\downarrow$ & PSNR$\uparrow$ & SSIM$\uparrow$ & \multicolumn{1}{c|}{LPIPS$\downarrow$} & PSNR$\uparrow$ & SSIM$\uparrow$ & LPIPS$\downarrow$ \\ 
 \midrule
T-NeRF\citep{Pumarola2021} & 13.163 & 0.709 & \multicolumn{1}{c|}{0.353} & 13.818 & 0.739 & 0.324 & 24.944 & 0.742 & \multicolumn{1}{c|}{0.336} & 22.242 & 0.700 & 0.363 \\
D-NeRF\citep{Pumarola2021} & 14.158 & 0.697 & \multicolumn{1}{c|}{0.352} & 14.660 & 0.737 & 0.312 & 25.380 & 0.766 & \multicolumn{1}{c|}{0.300} & 20.791 & 0.692 & 0.349 \\
TiNeuVox\citep{Fang2022} & 27.988 & 0.960 & \multicolumn{1}{c|}{0.063} & 19.612 & 0.940 & 0.073 & 29.982 & 0.864 & \multicolumn{1}{c|}{0.213} & 21.029 & 0.770 & 0.281 \\
T-NeRF$_{PINN}$ & 15.286 & 0.794 & \multicolumn{1}{c|}{0.293} & 16.189 & 0.835 & 0.230 & 16.250 & 0.441 & \multicolumn{1}{c|}{0.638} & 17.290 & 0.477 & 0.618 \\
HexPlane$_{PINN}$ & 27.042 & 0.958 & \multicolumn{1}{c|}{0.057} & 21.419 & 0.946 & 0.067 & 25.215 & 0.763 & \multicolumn{1}{c|}{0.389} & 23.091 & 0.742 & 0.401 \\
NSFF\citep{Li2021c} & - & - & \multicolumn{1}{c|}{-} & - & - & - & 29.365 & 0.829 & \multicolumn{1}{c|}{0.278} & 24.163 & 0.795 & 0.289 \\
NVFi\citep{Li2023c}  & 29.027 & 0.970 & \multicolumn{1}{c|}{0.039} & 27.594 & 0.972 & 0.036 & \underline{30.675} & 0.877 & \multicolumn{1}{c|}{0.211} & 29.745 & 0.876 & 0.204 \\
DefGS\citep{Yang2024} & \underline{37.865} & \underline{0.994} & \multicolumn{1}{c|}{\underline{0.007}} & 19.849 & 0.949 & 0.045 & 29.926 & \underline{0.916} & \multicolumn{1}{c|}{\underline{0.130}} & 21.380 & 0.819 & 0.188 \\ 
DefGS$_{nvfi}$ & 37.316 & \underline{0.994} & \multicolumn{1}{c|}{0.008} & 28.749 & \underline{0.984} & \underline{0.013} & 30.170 & 0.915 & \multicolumn{1}{c|}{0.133} & \underline{31.096} & \underline{0.945} & \underline{0.077}  \\ 
4DGS\citep{Wu2024} & 37.285 & 0.986 & 0.020 & 20.354 & 0.950 & 0.052 & 29.381 & 0.889 & 0.212 & 21.107 & 0.793 & 0.274 \\[+0.1em] 
\midrule 
\textbf{\nickname{}$_{4dgs}$ (Ours)} & 35.676 & 0.985 & 0.021 & \underline{30.327} & 0.983 & 0.019 & 28.804 & 0.862 & 0.242 & 30.607 & 0.896 & 0.209 \\
\textbf{\nickname{} (Ours)} & \textbf{39.305} & \textbf{0.995} & \multicolumn{1}{c|}{\textbf{0.006}} & \textbf{31.597} & \textbf{0.987} & \textbf{0.009} & \textbf{32.088} & \textbf{0.929} & \multicolumn{1}{c|}{\textbf{0.093}} & \textbf{34.824} & \textbf{0.965} & \textbf{0.054} \\
\bottomrule
\end{tabular}
}
\label{tab:exp_extrapolation}
\vspace{-0.3cm}
\end{table*}

\begin{table*}[t]\tabcolsep=0.08cm 
\centering
\caption{Quantitative results of all methods for future frame extrapolation optionally with novel view interpolation on NVIDIA Dynamic Scene Dataset, Dynamic Multipart Dataset}
\resizebox{1.\linewidth}{!}{
\begin{tabular}{r|ccc|ccc|ccc|ccc}
\toprule
\multirow{3}{*}{} & \multicolumn{6}{c|}{NVIDIA Dynamic Scene Dataset} & \multicolumn{6}{c}{Dynamic Multipart Dataset}  \\ \cmidrule{2-13} 
& \multicolumn{3}{c|}{Interpolation} & \multicolumn{3}{c|}{Extrapolation} & \multicolumn{3}{c|}{Interpolation} & \multicolumn{3}{c|}{Extrapolation} \\ \cmidrule{2-13} 
& PSNR$\uparrow$ & SSIM$\uparrow$ & \multicolumn{1}{c|}{LPIPS$\downarrow$} & PSNR$\uparrow$ & SSIM$\uparrow$ & LPIPS$\downarrow$ & PSNR$\uparrow$ & SSIM$\uparrow$ & \multicolumn{1}{c|}{LPIPS$\downarrow$} & PSNR$\uparrow$ & SSIM$\uparrow$ & LPIPS$\downarrow$\\ 
\midrule
T-NeRF\citep{Pumarola2021} & 23.078 & 0.684 & \multicolumn{1}{c|}{0.355} & 21.120 & 0.707 & 0.358 & 9.833 & 0.567 & \multicolumn{1}{c|}{0.550} & 10.064 & 0.576 & 0.537 \\
D-NeRF\citep{Pumarola2021} & 22.827 & 0.711 & \multicolumn{1}{c|}{0.309} & 20.633 & 0.709 & 0.327 & 13.279 & 0.747 & \multicolumn{1}{c|}{0.378} & 13.344 & 0.767 & 0.340 \\
TiNeuVox\citep{Fang2022} & \textbf{28.304} & 0.868 & \multicolumn{1}{c|}{0.216} & 24.556 & 0.863 & 0.215 & 29.957 & 0.966 & \multicolumn{1}{c|}{0.067} & 20.804 & 0.923 & 0.090 \\
T-NeRF$_{PINN}$ & 18.443 & 0.597 & \multicolumn{1}{c|}{0.439} & 17.975 & 0.605 & 0.428 & - & - & \multicolumn{1}{c|}{-} & - & - & - \\
HexPlane$_{PINN}$ & 24.971 & 0.818 & \multicolumn{1}{c|}{0.281} & 24.473 & 0.818 & 0.279 & - & - & \multicolumn{1}{c|}{-} & - & - & - \\
NVFi\citep{Li2023c} & \underline{27.138} & 0.844 & \multicolumn{1}{c|}{0.231} & \underline{28.462} & 0.876 & 0.214 & 27.516 & 0.960 & \multicolumn{1}{c|}{0.052} & 25.235 & 0.955 & 0.046 \\
DefGS\citep{Yang2024} & 26.662 & \underline{0.893} & \multicolumn{1}{c|}{\underline{0.127}} & 24.240 & 0.895 & 0.140 & 34.635 & 0.990 & \multicolumn{1}{c|}{0.019} & 20.664 & 0.930 & 0.067 \\ 
DefGS$_{nvfi}$ & 26.972 & 0.890 & \multicolumn{1}{c|}{0.128} & 27.529 & \underline{0.927} & \underline{0.102} & 34.637 & 0.990 & \multicolumn{1}{c|}{0.018} & 28.455 & 0.979 & 0.017 \\ 
4DGS\citep{Wu2024} & 19.411 & 0.462 & 0.532 & 22.510 & 0.703 & 0.408 & \textbf{37.021} & \textbf{0.992} & \underline{0.014} & 20.564 & 0.935 & 0.067 \\[+0.1em] 
\midrule 
\textbf{\nickname{}$_{4dgs}$ (Ours)} & 18.618 & 0.432 & 0.553 & 22.554 & 0.721 & 0.390 & \underline{35.993} & \underline{0.991} & 0.016 & \underline{32.317} & \underline{0.985} & \underline{0.016} \\
\textbf{\nickname{} (Ours)} & 26.861 & \textbf{0.912} & \textbf{0.089} & \textbf{29.341} & \textbf{0.933} & \textbf{0.074} & 35.768 & \underline{0.991} & \textbf{0.011} & \textbf{33.481} & \textbf{0.990} & \textbf{0.007} \\ 
\bottomrule
\end{tabular}
}
\label{tab:exp_extrapolation_real_multipart}
\vspace{-0.3cm}
\end{table*}

\begin{figure*}[t!]
\centering
\includegraphics[width=1.0\linewidth]{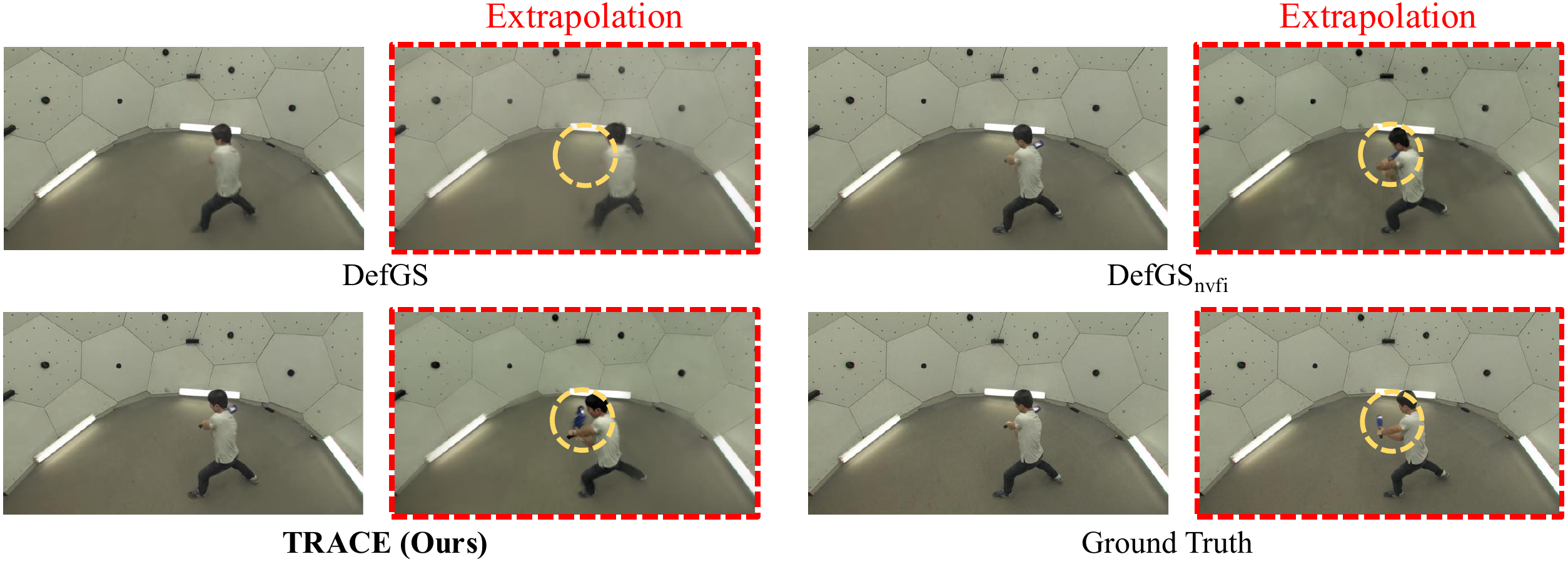}
\caption{Qualitative Results on Panoptic Sports Dataset.}
\label{fig:qualitative_pano}
\vspace{-0.4cm}
\end{figure*}

\subsection{Quantitative and qualitative results for continual learning on Panoptic Sports dataset}\label{sec:extrap_panoptic}

\begin{figure}[h]
\centering
\vspace{-0.8cm}
\includegraphics[width=1.0\linewidth]{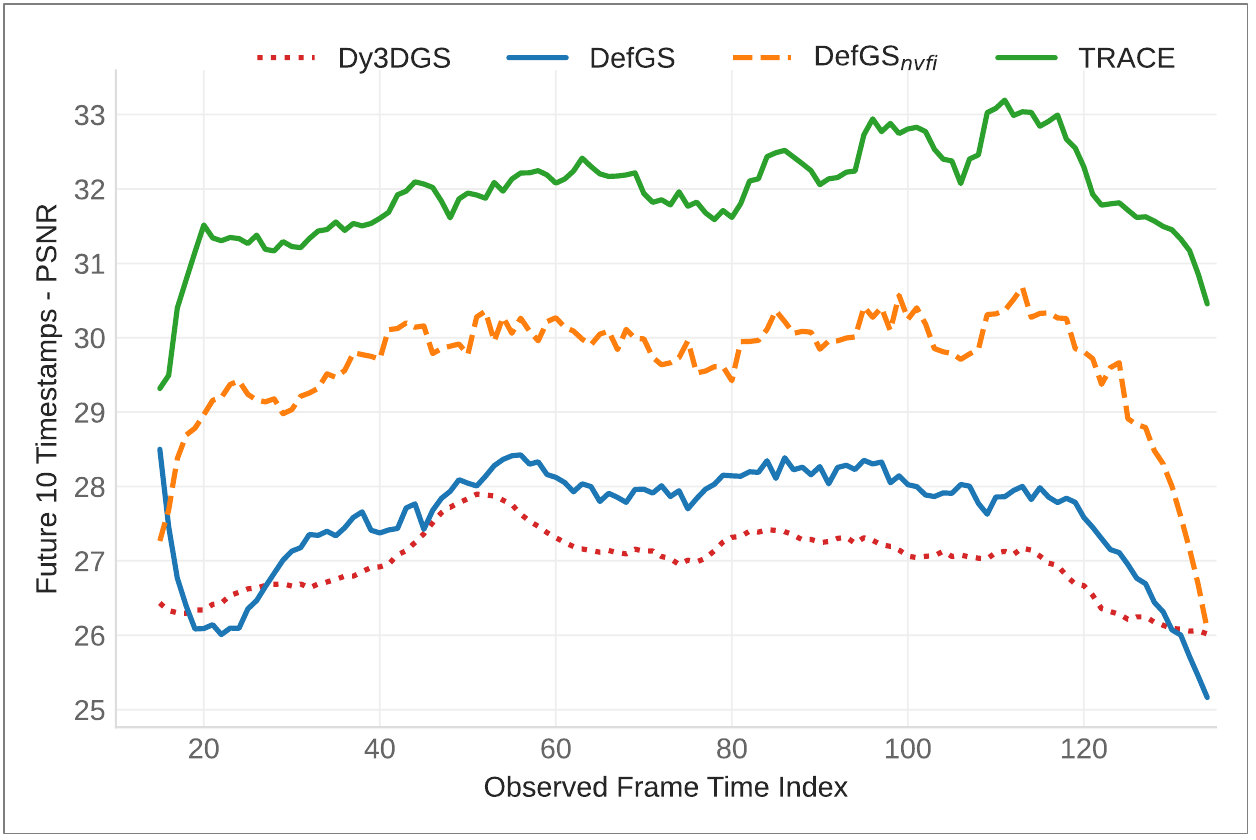}
\caption{Quantitative Results on Panoptic Sports Dataset.}
\label{fig:quantitative_pano}
\vspace{-0.4cm}
\end{figure}

We also conduct experiments on the Panoptic Sports Dataset. It has 6 realistic non-rigid scenes like human interactions, where each scene has 31 viewpoints and each viewpoint has 150 timestamps/frames in total. This dataset is naturally suitable for our continual learning setting and we evaluate our method as follows: 1) the first 15 timestamps are selected for initial training; 2) for every later round, we load one more timestamp into training set and evaluate future prediction on the future 10 timestamps; 3) we keep training till 135$^{th}$ timestamp. Quantitative results are presented in Figure \ref{fig:quantitative_pano} and qualitative results are shown in Figure \ref{fig:qualitative_pano}. We can see that our method succeeds in extremely challenging non-rigid scenes with sudden motions and non-constant forces. By design, our method is also flexible to advanced hierarchical deformation fields, and Gaussians are traced in a Lagrangian representation, thus mass being preserved in nature. 

\subsection{Implementation Details for the Segmentation of DefGS / DefGS$_{nvfi}$}
\label{app:ogc_detail}

Given a well-trained DefGS or DefGS$_{nvfi}$ model with $N$ canonical Gaussian kernels, we first assign learnable per-Gaussian object codes $\boldsymbol{O} \in {(0, 1)}^{N \times K}$ to all Gaussian kernels, where $K$ is the maximum number of objects that is expected to appear in the scene.

After that, we query the position displacements for Gaussians kernels from the well-trained deformation field at time $0$ and $t$ respectively, thus obtaining the Gaussians $\boldsymbol{P}_0$ at time $0$ and $\boldsymbol{P}_t$ at time $t$. Then the per-Gaussian scene flows $\boldsymbol{M}_t$ from time $0$ to $t$ is calculated as $\boldsymbol{M}_t = \boldsymbol{P}_t - \boldsymbol{P}_0$.

Lastly, two losses proposed in OGC \cite{Song2022} are computed on the learnable object codes. \textbf{1) Dynamic rigid consistency:} For the $k^{th}$ object, we first retrieve its (soft) binary mask $\boldsymbol{O}^k$, and feed the tuple $\{\boldsymbol{P}_0, \boldsymbol{P}_t, \boldsymbol{O}^k\}$ into the weighted-Kabsch algorithm to estimate its transformation matrix $\boldsymbol{T}_k \in \mathbb{R}^{4\times 4}$ belonging to $SE(3)$ group. Then the dynamic loss is computed as:
\begin{equation*}
    \ell_{dynamic} = \frac{1}{N} \sum_{\boldsymbol{p} \in \boldsymbol{P}_0} \Big\| \Big(\sum_{k=1}^K {o}_{\boldsymbol{p}}^{k} \cdot (\boldsymbol{T}_k \circ \boldsymbol{p}) \Big) - (\boldsymbol{p} + \boldsymbol{m}_t) \Big\|_2
\end{equation*}
where ${o}_{\boldsymbol{p}}^{k}$ represents the probability of being assigned to the $k^{th}$ object for a specific point $\boldsymbol{p}$, and $\boldsymbol{m}_t \in \mathbb{R}^{3}$ represents the motion vector of $\boldsymbol{p}$ from time $0$ to $t$. The operation $\circ$ applies the rigid transformation to the point. This loss aims to discriminate objects with different motions. \textbf{2) Spatial smoothness:} For each point $\boldsymbol{p}$ in $\boldsymbol{P}_0$, we first search $H$ nearest neighboring points. Then the smoothness loss is defined as:
\begin{equation}
    \ell_{smooth} = \frac{1}{N} \sum_{\boldsymbol{p} \in \boldsymbol{P}_0} \Big( \frac{1}{H}\sum_{h=1}^H \| \boldsymbol{o}_{\boldsymbol{p}} - \boldsymbol{o}_{\boldsymbol{p}_h} \|_1 \Big)
\end{equation}
where $\boldsymbol{o}_{\boldsymbol{p}} \in {(0, 1)}^{K}$ represents the object assignment of the center point $\boldsymbol{p}$, and $\boldsymbol{o}_{\boldsymbol{p}_h} \in {(0, 1)}^{K}$ represents the object assignment of its $h^{th}$ neighboring point. This loss aims to avoid the over-segmentation issues. More details are provided in \cite{Song2022}.

In our experiments for DefGS and DefGS$_{nvfi}$, the maximum number of predicted objects $K$ is set to be 8. A softmax activation is applied on per-Gaussian object codes. During optimization, we adopt the Adam optimizer with a learning rate of 0.01 and optimize object codes for 1000 iterations until convergence.

\subsection{Segmentation on Real-world Scene}\label{sec:real-seg}

\begin{figure}[h]\vspace{-0.2cm}
\centering
\caption{Qualitative results of segmentation.}
\includegraphics[width=0.8\linewidth]{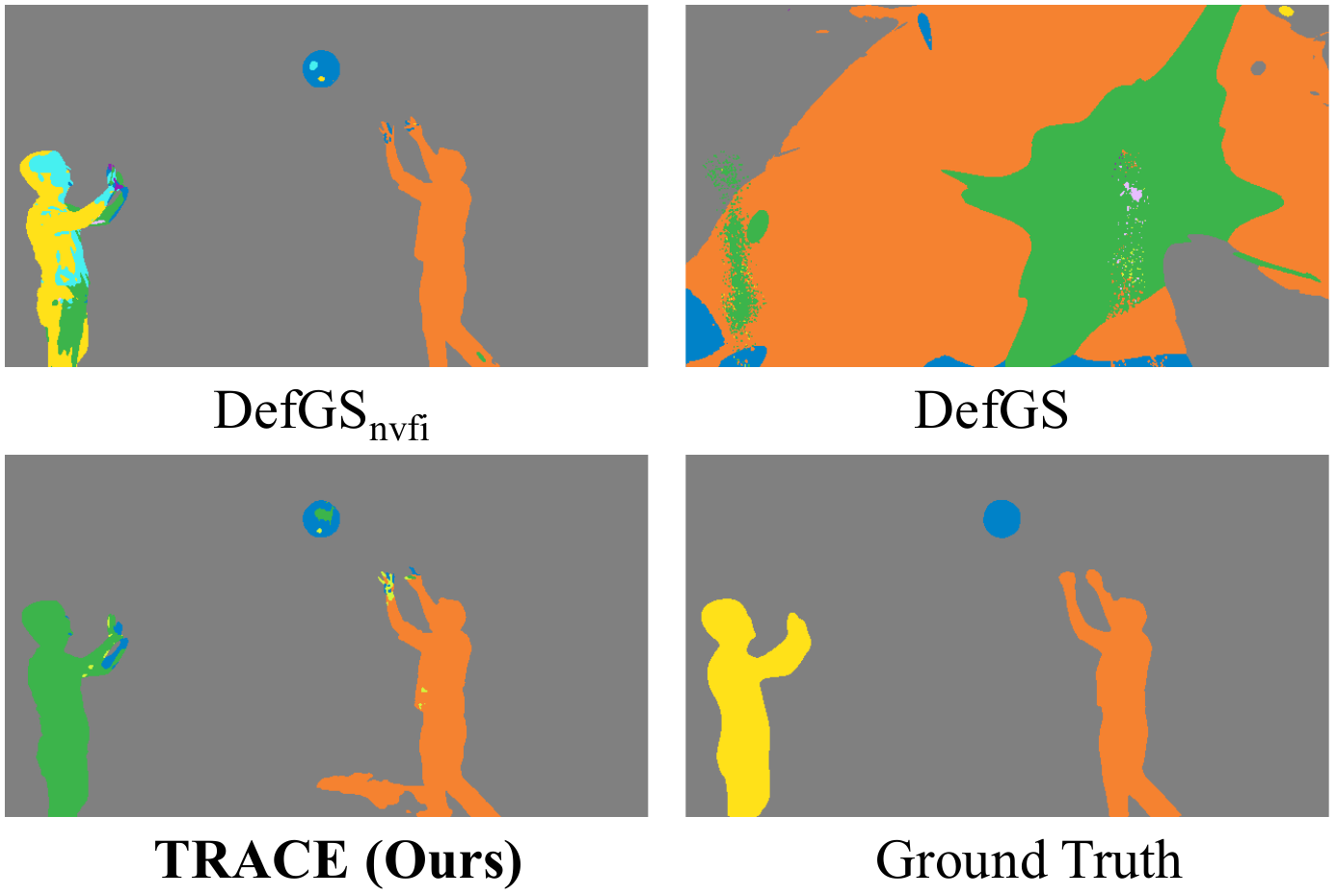}
\label{fig:rebuttal_pano_seg}
\vspace{-0.8cm}
\end{figure}

We select the basketball scene from Panoptic Sports Dataset, and use manual prompts on Track-Anything to get pseudo ground truth. We render 2D masks on all $31\times150=4650$ views for evaluation. Results are shown in Table \ref{tab:rebuttal-seg} and Figure \ref{fig:rebuttal_pano_seg}.
\setlength{\abovecaptionskip}{0 pt}
\begin{table}[ht]\vspace{-0.4cm} 
\tabcolsep=0.2cm 
\caption{Segmentation results on basketball scene of Panoptic Sports Dataset.}
\label{tab:rebuttal-seg}
\footnotesize
\resizebox{1.\linewidth}{!}{
\begin{tabular}{l|cccccc}
\hline
            & AP$\uparrow$    & PQ$\uparrow$    & F1$\uparrow$    & Pre$\uparrow$    & Rec$\uparrow$    & mIoU$\uparrow$  \\ 
\hline
DefGS & 5.64 & 15.30 & 20.51 & 15.53 & 30.23 & 32.05 \\
DefGS$_{nvfi}$ & 43.52 & 53.60 & 64.02 & 52.95 & 80.93 & \textbf{68.25} \\ \hline
\textbf{\nickname{} (Ours)} & \textbf{53.40} & \textbf{59.46} & \textbf{72.06} & \textbf{64.64} & \textbf{81.40} & 66.19 \\\hline

\end{tabular}
}
\vspace{-0.4cm}
\end{table}

\begin{figure*}[t]
    \centering
    \includegraphics[width=0.9\linewidth]{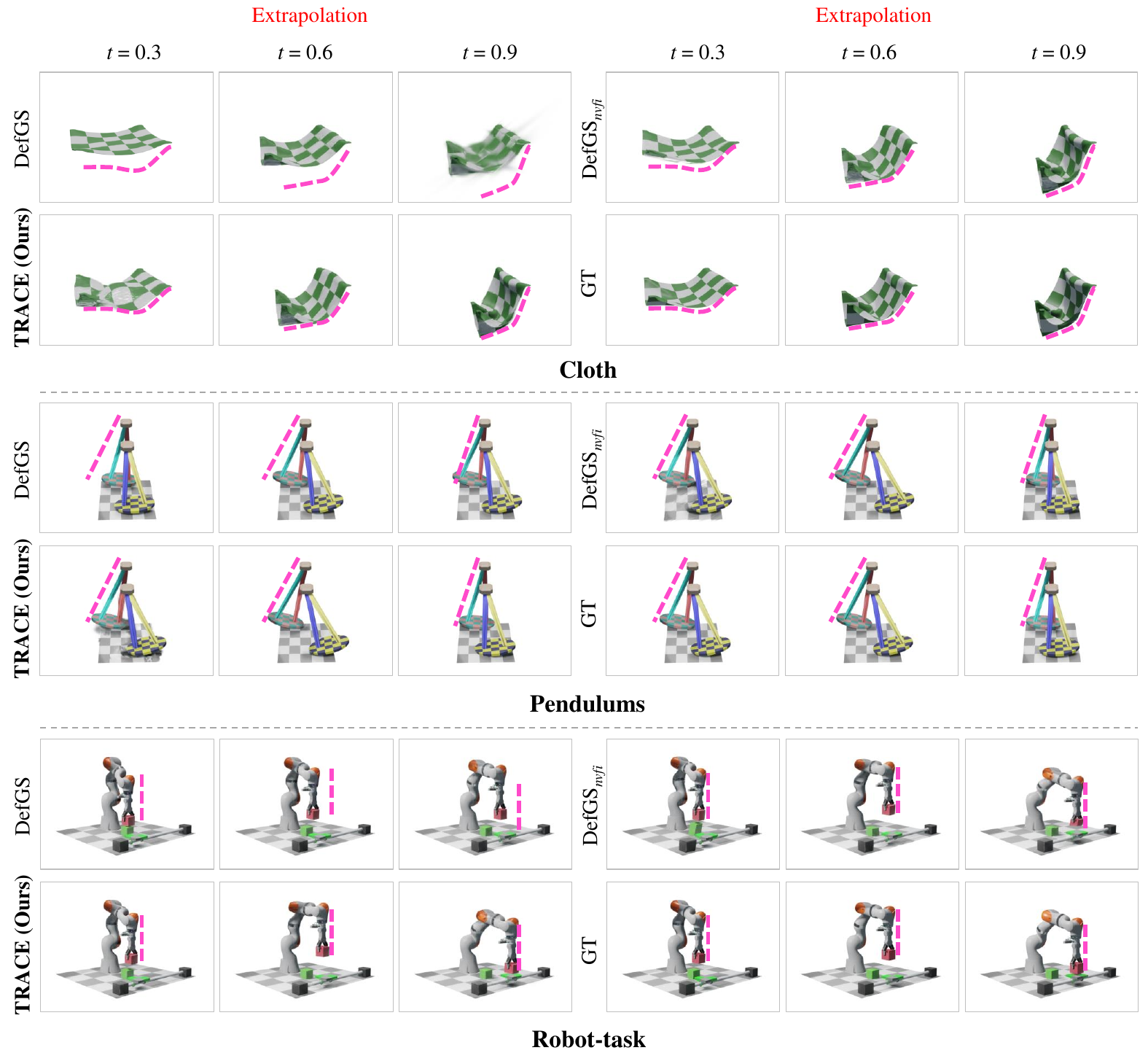}
    \caption{Qualitative \textcolor{red}{extrapolation} results for continual learning on Particle NeRF dataset.}
    \label{fig:particle_nerf}
\end{figure*}

\subsection{Details of Particle NeRF Dataset and Additional Results for Continual Learning}
\label{app:incremental}

We first introduce details for Particle NeRF datasets. This dataset comprises 6 scenes, and each has 40 views. We reserve 36 views for training and 4 views for novel view evaluation. The details for motions and time splits for each scene are listed following:
\begin{itemize}
    \item \textbf{cloth:} This scene includes a piece of square cloth folded by external forces. 
    \item \textbf{wheel:}  This scene includes a rolling wheel with a constant angular velocity.
    \item \textbf{spring:} This scene includes a box tied onto a spring, and undergoing a harmonic oscillation motion.
    \item \textbf{robot:}  This scene shows a robot arm rotating its poses.
    \item \textbf{robot-task:} This scene shows a robot arm putting a box onto a sliding platform.
    \item \textbf{pendulums:} This scene shows two pendulums and each of them undergoes harmonic oscillation motion. 
\end{itemize}

For the first three scenes, we define the 140-th frame as virtual time $1.0$, and for the latter three scenes we define the 70-th frame as virtual time $1.0$. 

Figure \ref{fig:particle_nerf} shows the qualitative future frame extrapolation results. Table \ref{tab:exp_incremental_particle_interp} shows the quantitative past-time interpolation results.

\begin{table}[ht]\tabcolsep=0.25cm  \vspace{-0.3cm}
\centering
\caption{Quantitative \textcolor{red}{interpolation} results (PSNR) of continual learning.}
\resizebox{1\linewidth}{!}{
\begin{tabular}{r|ccccc|c}
\toprule
Observed Time & 0.15 & 0.30 & 0.45 & 0.60 & 0.75 & \multirow{2}*{Average} \\ 
Extrap Till & 0.30 & 0.45 & 0.60 & 0.75 & 0.90 &  \\ 
 \midrule
DefGS\citep{Yang2024} & 38.665 & 37.446 & 36.730 & 36.229 & 35.820 & 36.978\\
DefGS$_{nvfi}$ & 37.984 & 36.499 & 35.233 & 35.168 & 34.896 & 35.956 \\
\textbf{\nickname{} (Ours)} & 35.213 & 34.705 & 33.164 & 33.567 & 33.414 & 34.013 \\
\bottomrule
\end{tabular}
}
\label{tab:exp_incremental_particle_interp}
\vspace{-0.5cm}
\end{table}

\subsection{Additional details of our new dataset}
\label{sec:details_datasets}

\paragraph{Dynamic Multipart dataset:} This dataset comprises 4 distinct objects \footnote{All objects are purchased from SketchFab, licensed under the SketchFab Standard License: https://sketchfab.com/licenses, and are all allowed for AI generation model usage}, including a variety of challenging motions. Details of the 4 dynamic objects are:
\begin{itemize}
    \item \textbf{Foldingchair:} A folingchair is given. This chair is composed of three parts. The whole motion is unfolding this chair, so all three parts are undergoing different rotating motions.  
    \item \textbf{Hypoerbolic Slot:} This is an extremely hard case, where a stick is rotating through a hypoerbolic slot. Note that, only the stick in this hyperbolic shape is dynamic, this introduces more challenges in motion extrapolation.  
    \item \textbf{Satellite:} This object is a satellite with two wing doors opening and one main door opening, all rotating in different directions. 
    \item \textbf{Stove:} A home stove is given. The motion is mainly closing its top cover plate. 
\end{itemize}

\subsection{Additional Quantitative Results for Ablation Study for the main context}\label{sec:app_res_ablation}

Here we show the total results for the ablation study in Table \ref{tab:ablation_all}, both for interpolation and extrapolation.

\setlength{\abovecaptionskip}{0 pt}
\begin{table}[ht] \tabcolsep=0.035cm  \vspace{-0.2cm}
\caption{Ablation studies on three datasets.}
\label{tab:ablation_all}
\footnotesize
\resizebox{1\linewidth}{!}{
\begin{tabular}{lcccc|ccc|ccc|ccc}
\toprule
\multicolumn{14}{c}{Interpolation} \\
\midrule
 &  &  &  &  & \multicolumn{3}{c|}{Dynamic Multipart} & \multicolumn{3}{c|}{Dynamic Object} & \multicolumn{3}{c}{Dynamic Indoor Scene} \\ \cmidrule{6-14} 
  & order & $f_{defo}$ & physcis & equiv & PSNR$\uparrow$ & SSIM$\uparrow$ & LPIPS$\downarrow$ & PSNR$\uparrow$ & SSIM$\uparrow$ & LPIPS$\downarrow$ & PSNR$\uparrow$ & SSIM$\uparrow$ & LPIPS$\downarrow$ \\
\midrule
(1) $\delta t$ & 2 & \Y & \Y & \Y & 35.370 & 0.991 & 0.018 & \textbf{39.305} & \textbf{0.995} & 0.006 & 32.184 & 0.928 & 0.121 \\ 
(1) $2\delta t$ & 2 & \Y & \Y & \Y & \textbf{35.768} & \textbf{0.991} & \textbf{0.011} & 38.518 & 0.995 & \textbf{0.005} & 32.088 & \textbf{0.929} & \textbf{0.093} \\
(1) $3\delta t$ & 2 & \Y & \Y & \Y & 35.666 & 0.991 & 0.011 & 37.718 & 0.994 & 0.005 & 32.044 & 0.929 & 0.094 \\
\midrule
(2) $2\delta t$ & 1 & \Y & \Y & \Y & 35.369 & 0.991 & 0.018 & 37.859 & 0.994 & 0.006 & \textbf{32.250} & 0.927 & 0.122 \\
(2) $2\delta t$ & 3 & \Y & \Y & \Y & 35.832 & 0.991 & 0.011 & 38.669 & 0.995 & 0.004 & 32.111 & 0.929 & 0.093 \\
\midrule
(3) $2\delta t$ & 2 & \N & \Y & \Y & 18.906 & 0.895 & 0.133 & - & - & - & - & - & - \\
(4) $2\delta t$ & 2 & \Y & \N & \Y & 35.526 & 0.992 & 0.011 & - & - & - & - & - & - \\
(5) $2\delta t$ & 2 & \Y & \Y & \N & 35.225 & 0.991 & 0.011 & - & - & - & - & - & - \\ 
\midrule
\multicolumn{14}{c}{Extrapolation} \\
\midrule
&  &  &  &  & \multicolumn{3}{c|}{Dynamic Multipart} & \multicolumn{3}{c|}{Dynamic Object} & \multicolumn{3}{c}{Dynamic Indoor Scene} \\ \cmidrule{6-14} 
  & order & $f_{defo}$ & physcis & equiv & PSNR$\uparrow$ & SSIM$\uparrow$ & LPIPS$\downarrow$ & PSNR$\uparrow$ & SSIM$\uparrow$ & LPIPS$\downarrow$ & PSNR$\uparrow$ & SSIM$\uparrow$ & LPIPS$\downarrow$ \\
\midrule
(1) $\delta t$ & 2 & \Y & \Y & \Y & 32.852 & 0.989 & 0.010 & \textbf{31.597} & 0.987 & 0.009 & 33.831 & 0.958 & 0.078 \\ 
(1) $2\delta t$ & 2 & \Y & \Y & \Y & \textbf{33.481} & \textbf{0.990} & \textbf{0.007} & 31.511 & \textbf{0.988} & \textbf{0.006} & \textbf{34.824} & \textbf{0.965} & \textbf{0.054} \\
(1) $3\delta t$ & 2 & \Y & \Y & \Y & 33.003 & 0.989 & 0.008 & 29.787 & 0.983 & 0.011 & 34.778 & \textbf{0.965} & \textbf{0.054} \\
\midrule
(2) $2\delta t$ & 1 & \Y & \Y & \Y & 33.125 & 0.989 & 0.010 & 28.522 & 0.984 & 0.012 & 34.576 & 0.963 & 0.076 \\
(2) $2\delta t$ & 3 & \Y & \Y & \Y & 33.312 & 0.989 & 0.008 & 31.044 & 0.987 & 0.007 & 34.086 & 0.961 & 0.056\\
\midrule
(3) $2\delta t$ & 2 & \N & \Y & \Y & 19.206 & 0.907 & 0.120 & - & - & - & - & - & - \\
(4) $2\delta t$ & 2 & \Y & \N & \Y & 29.602 & 0.983 & 0.012 & - & - & - & - & - & - \\
(5) $2\delta t$ & 2 & \Y & \Y & \N & 29.986 & 0.985 & 0.012 & - & - & - & - & - & - \\ 
\bottomrule
\end{tabular}
}
\vspace{-0.3cm}
\end{table}

\subsection{Analysis on Performance Change along Prediction Time Span}\label{sec:performance-decrease}
\begin{figure}[h] 
\centering
\vspace{-0.3cm}
\caption{The top row shows extrapolation results over future timestamps and the bottom shows qualitative results of our extrapolation.}
\includegraphics[width=1.0\linewidth]{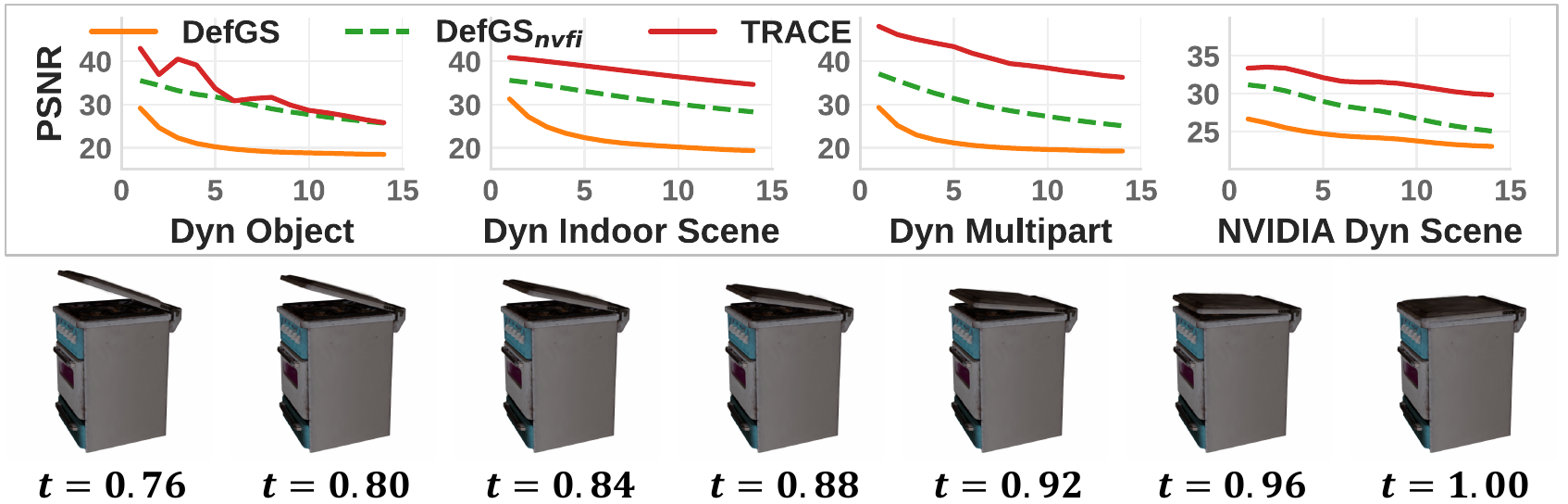}
\label{fig:rebuttal_14frames}
\vspace{-0.7cm}
\end{figure}

We present the extrapolation results for per future timestamp in Figure \ref{fig:rebuttal_14frames}. We can see that, though our method is clearly better than baselines, the performance of all methods decreases as the extrapolation goes further. To alleviate this issue, we advocate incorporating new observations in a continual learning manner. 

\subsection{Analysis on Performance against Motion Complexity}\label{sec:complexity}
\begin{figure}[h]\vspace{-0.1cm}
\centering
\vspace{-0.3cm}
\caption{Extrapolation over different scene motion complexities.}
\includegraphics[width=1.0\linewidth]{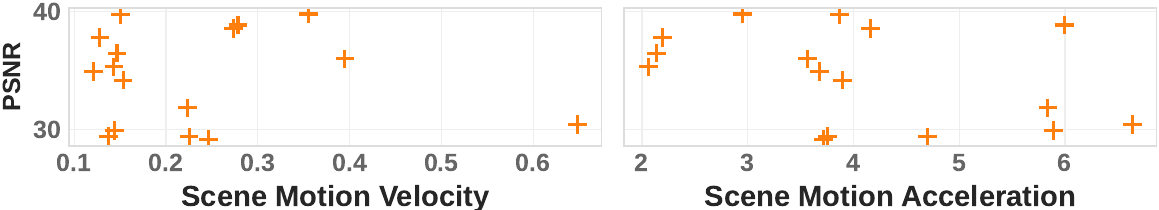}
\label{fig:rebuttal_complexity}
\vspace{-0.8cm}
\end{figure}

For each scene, we calculate its scene motion complexity as follows: 1) we uniformly sample 40K points on each image of the first timestamp; 2) we use BootsTAPIR\cite{doersch2024bootstap} to track these points over all timestamps and static points are removed; 3) we calculate optical flows for dynamic points between every two adjacent timestamps; 4) the optical flows are divided by the time interval and then normalized by image size, getting normalized image motion velocity which will be averaged out on the whole scene to obtain \textit{Scene Motion Velocity}; 5) we further calculate the difference between two adjacent normalized image motion velocities, getting acceleration which will be averaged out on the whole scene to obtain \textit{Scene Motion Acceleration}. Figure \ref{fig:rebuttal_complexity} shows the relationship where each cross represents one scene out of our 4 datasets. We can see that we achieve similar results over different levels of motion complexity, highlighting the generality of our method.    

\subsection{Additional Quantitative Results on Dynamic Object Dataset}\label{sec:app_res_dynObj}
\begin{figure*}[t]
    \centering
    \includegraphics[width=0.85\linewidth]{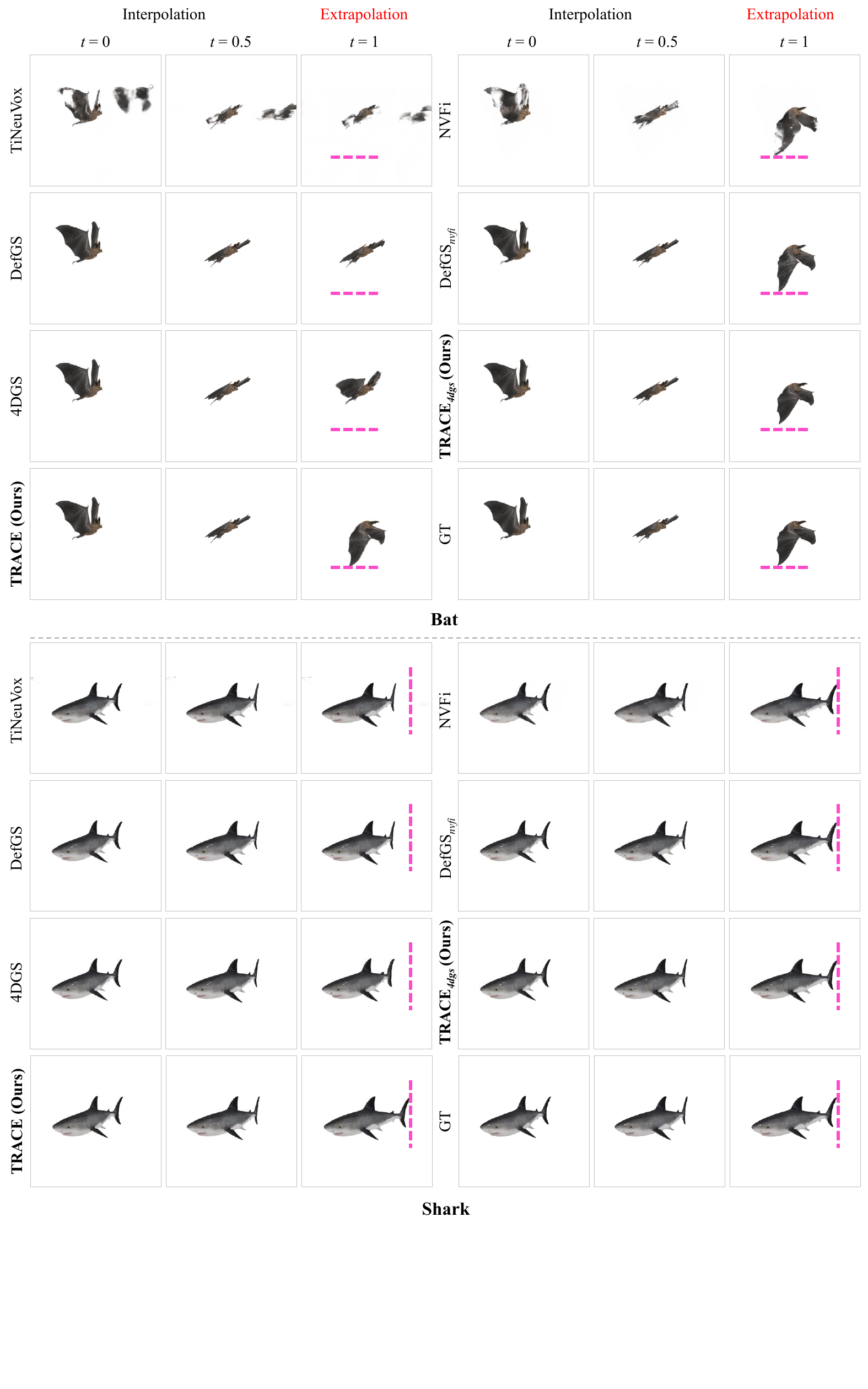}
    \vspace*{-25mm}
    \caption{Qualitative results of RGB view synthesis for interpolation and \textcolor{red}{extrapolation} tasks on Dynamic Object dataset.}
    \label{fig:app_qual_res_object1}
\end{figure*}

\begin{figure*}[t]
    \centering
    \includegraphics[width=0.85\linewidth]{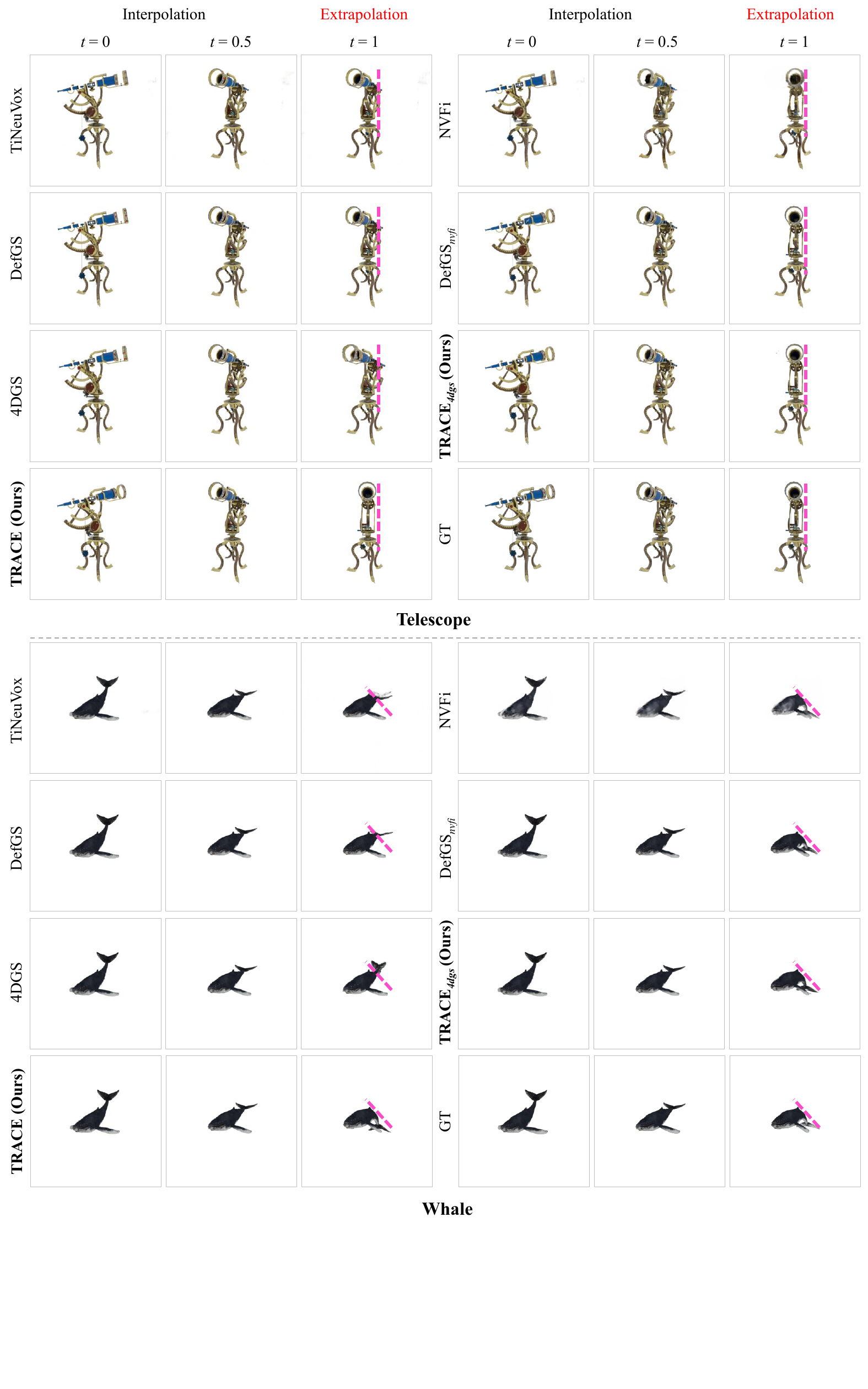}
    \vspace*{-25mm}
    \caption{Qualitative results of RGB view synthesis for interpolation and \textcolor{red}{extrapolation} tasks on Dynamic Object dataset.}
    \label{fig:app_qual_res_object2}
\end{figure*}

Here we show the total per-scene results on Dynamic Object Dataset in Table \ref{tab:object_all} and qualitative results in Figures \ref{fig:app_qual_res_object1}, \ref{fig:app_qual_res_object2}\&\ref{fig:app_qual_res_indoor1}, both for interpolation and extrapolation.

\begin{table*}[t]\tabcolsep=0.1cm 
\footnotesize
\centering
\caption{Per-scene quantitative results on Dynamic Object dataset.}
\resizebox{1.0\linewidth}{!}{
\begin{tabular}{r|cccccc|cccccc}
\toprule
 & \multicolumn{6}{c|}{Falling Ball} & \multicolumn{6}{c}{Bat} \\ \cmidrule{2-13} 
Methods & \multicolumn{3}{c}{Interpolation} & \multicolumn{3}{c|}{Extrapolation} & \multicolumn{3}{c}{Interpolation} & \multicolumn{3}{c}{Extrapolation} \\ \cmidrule{2-13}
 & PSNR$\uparrow$ & SSIM$\uparrow$ & LPIPS$\downarrow$ & PSNR$\uparrow$ & SSIM$\uparrow$ & LPIPS$\downarrow$ & PSNR$\uparrow$ & SSIM$\uparrow$ & LPIPS$\downarrow$ & PSNR$\uparrow$ & SSIM$\uparrow$ & LPIPS$\downarrow$ \\ \midrule
T-NeRF\citep{Pumarola2021} & 14.921 & 0.782 & 0.326 & 15.418 & 0.793 & 0.308 & 13.070 & 0.836 & 0.234 & 13.897 & 0.834 & 0.230 \\
D-NeRF\citep{Pumarola2021} & 15.548 & 0.665 & 0.435 & 15.116 & 0.644 & 0.427 & 14.087 & 0.845 & 0.212 & 15.406 & 0.887 & 0.175 \\
TiNeuVox\citep{Fang2022} & 35.458 & 0.974 & 0.052 & 20.242 & 0.959 & 0.067 & 16.080 & 0.908 & 0.108 & 16.952 & 0.930 & 0.115 \\
T-NeRF$_{PINN}$ & 17.687 & 0.775 & 0.368 & 17.857 & 0.829 & 0.265 & 16.412 & 0.903 & 0.197 & 18.983 & 0.930 & 0.132 \\
HexPlane$_{PINN}$ & 32.144 & 0.965 & 0.065 & 20.762 & 0.951 & 0.081 & 23.399 & 0.958 & 0.057 & 21.144 & 0.951 & 0.064 \\ 
NVFi\citep{Li2023c}  & 35.826 & 0.978 & 0.041 & 31.369 & 0.978 & 0.041 & 23.325 & 0.964 & 0.046 & 25.015 & 0.968 & 0.042 \\ 
DefGS\citep{Yang2024} & 37.535 & 0.995 & 0.009 & 20.442 & 0.976 & 0.033 & 38.750 & \textbf{0.997} & 0.004 & 17.063 & 0.936 & 0.072 \\ 
DefGS$_{nvfi}$ & 38.606 & 0.996 & 0.010 & 24.873 & 0.985 & 0.015 & 38.075 & \textbf{0.997} & 0.004 & \textbf{28.950} & \textbf{0.980} & \textbf{0.015} \\ 
\midrule
\textbf{\nickname{}$_{4dgs}$ (Ours)} & 27.481 & 0.943 & 0.083 & 29.951 & 0.974 & 0.050 & 38.354 & \textbf{0.997} & 0.004 & 28.093 & \textbf{0.980} & \textbf{0.015} \\
\textbf{\nickname{} (Ours)} & \textbf{41.394} & \textbf{0.998} & \textbf{0.005} & \textbf{42.713} & \textbf{0.998} & \textbf{0.004} & \textbf{39.384} & \textbf{0.997} & \textbf{0.003} & 26.449 & 0.979 & \textbf{0.015} \\ \midrule

 & \multicolumn{6}{c|}{Fan} & \multicolumn{6}{c}{Telescope} \\ \cmidrule{2-13} 
Methods & \multicolumn{3}{c}{Interpolation} & \multicolumn{3}{c|}{Extrapolation} & \multicolumn{3}{c}{Interpolation} & \multicolumn{3}{c}{Extrapolation} \\ \cmidrule{2-13}
 & PSNR$\uparrow$ & SSIM$\uparrow$ & LPIPS$\downarrow$ & PSNR$\uparrow$ & SSIM$\uparrow$ & LPIPS$\downarrow$ & PSNR$\uparrow$ & SSIM$\uparrow$ & LPIPS$\downarrow$ & PSNR$\uparrow$ & SSIM$\uparrow$ & LPIPS$\downarrow$ \\ \midrule
T-NeRF\cite{Pumarola2021} & 8.001 & 0.308 & 0.646 & 8.494 & 0.392 & 0.593 & 13.031 & 0.615 & 0.472 & 13.892 & 0.670 & 0.417 \\
D-NeRF\cite{Pumarola2021} & 7.915 & 0.262 & 0.690 & 8.624 & 0.370 & 0.623 & 13.295 & 0.609 & 0.469 & 14.967 & 0.700 & 0.385 \\
TiNeuVox\cite{Fang2022} & 24.088 & 0.930 & 0.104 & 20.932 & 0.935 & 0.078 & 31.666 & 0.982 & 0.041 & 20.456 & 0.921 & 0.067 \\
T-NeRF$_{PINN}$ & 9.233 & 0.541 & 0.508 & 9.828 & 0.606 & 0.443 & 14.293 & 0.739 & 0.366 & 15.752 & 0.804 & 0.298 \\
HexPlane$_{PINN}$ & 22.822 & 0.921 & 0.079 & 19.724 & 0.919 & 0.080 & 25.381 & 0.948 & 0.066 & 23.165 & 0.932 & 0.074 \\ 
NVFi\cite{Li2023c}  & 25.213 & 0.948 & 0.049 & 27.172 & 0.963 & 0.037 & 26.487 & 0.959 & 0.048 & 27.101 & 0.963 & 0.046 \\ 
DefGS\cite{Yang2024} & 35.858 & \textbf{0.985} & 0.017 & 20.932 & 0.948 & 0.038 & 37.502 & 0.996 & \textbf{0.003} & 20.684 & 0.927 & 0.048 \\ 
DefGS$_{nvfi}$ & 35.217 & 0.984 & 0.019 & 26.648 & 0.972 & 0.023 & 37.568 & 0.996 & \textbf{0.003} & \textbf{34.096} & \textbf{0.994} & \textbf{0.005} \\ 
\midrule
\textbf{\nickname{}$_{4dgs}$ (Ours)} & \textbf{36.052} & \textbf{0.985} & 0.019 & 34.622 & 0.988 & 0.012 & 37.668 & 0.994 & 0.006 & 33.338 & 0.989 & 0.010 \\
\textbf{\nickname{} (Ours)} & 35.969 & \textbf{0.985} & \textbf{0.015} & \textbf{34.964} & \textbf{0.990} & \textbf{0.008} & \textbf{40.441} & \textbf{0.997} & \textbf{0.003} & 31.145 & 0.986 & 0.006 \\ 
\midrule

 & \multicolumn{6}{c|}{Shark} & \multicolumn{6}{c}{Whale} \\ \cmidrule{2-13} 
Methods & \multicolumn{3}{c}{Interpolation} & \multicolumn{3}{c|}{Extrapolation} & \multicolumn{3}{c}{Interpolation} & \multicolumn{3}{c}{Extrapolation} \\ \cmidrule{2-13}
 & PSNR$\uparrow$ & SSIM$\uparrow$ & LPIPS$\downarrow$ & PSNR$\uparrow$ & SSIM$\uparrow$ & LPIPS$\downarrow$ & PSNR$\uparrow$ & SSIM$\uparrow$ & LPIPS$\downarrow$ & PSNR$\uparrow$ & SSIM$\uparrow$ & LPIPS$\downarrow$ \\ \midrule
T-NeRF\cite{Pumarola2021} & 13.813 & 0.853 & 0.223 & 15.325 & 0.882 & 0.193 & 16.141 & 0.860 & 0.212 & 15.880 & 0.860 & 0.203 \\
D-NeRF\cite{Pumarola2021} & 17.727 & 0.903 & 0.150 & 19.078 & 0.936 & 0.092 & 16.373 & 0.898 & 0.154 & 14.771 & 0.883 & 0.171 \\
TiNeuVox\cite{Fang2022} & 23.178 & 0.971 & 0.059 & 19.463 & 0.950 & 0.050 & 37.455 & 0.994 & 0.016 & 19.624 & 0.943 & 0.063 \\
T-NeRF$_{PINN}$ & 17.315 & 0.878 & 0.177 & 18.739 & 0.921 & 0.115 & 16.778 & 0.927 & 0.141 & 15.974 & 0.919 & 0.127 \\
HexPlane$_{PINN}$ & 28.874 & 0.976 & 0.040 & 22.330 & 0.961 & 0.047 & 29.634 & 0.981 & 0.035 & 21.391 & 0.961 & 0.053 \\ 
NVFi\cite{Li2023c}  & 32.072 & 0.984 & 0.024 & 28.874 & 0.982 & 0.021 & 31.240 & 0.986 & 0.025 & 26.032 & 0.978 & 0.029 \\ 
DefGS\cite{Yang2024} & 37.802 & 0.994 & 0.006 & 19.924 & 0.957 & 0.034 & 39.740 & \textbf{0.997} & \textbf{0.004} & 20.048 & 0.951 & 0.046 \\ 
DefGS$_{nvfi}$ & 37.327 & 0.994 & 0.006 & \textbf{29.240} & \textbf{0.987} & \textbf{0.007} & 37.101 & 0.996 & 0.005 & 28.686 & 0.986 & \textbf{0.012} \\ 
\midrule
\textbf{\nickname{}$_{4dgs}$ (Ours)} & 39.485 & 0.996 & 0.006 & 26.997 & 0.980 & 0.010 & 35.017 & 0.995 & 0.008 & \textbf{28.961} & \textbf{0.987} & 0.017 \\
\textbf{\nickname{} (Ours)} & \textbf{40.537} & \textbf{0.997} & \textbf{0.005} & 27.280 & 0.982 & 0.008 & \textbf{38.104} & \textbf{0.997} & \textbf{0.004} & 27.030 & 0.985 & \textbf{0.012} \\ 
\bottomrule
\end{tabular}
}
\label{tab:object_all}
\vspace{-0.1cm}
\end{table*}

\subsection{Additional Quantitative Results on Dynamic Indoor Scene Dataset}\label{sec:app_res_indoorScene}
\begin{figure*}[t]
    \centering
    \includegraphics[width=0.85\linewidth]{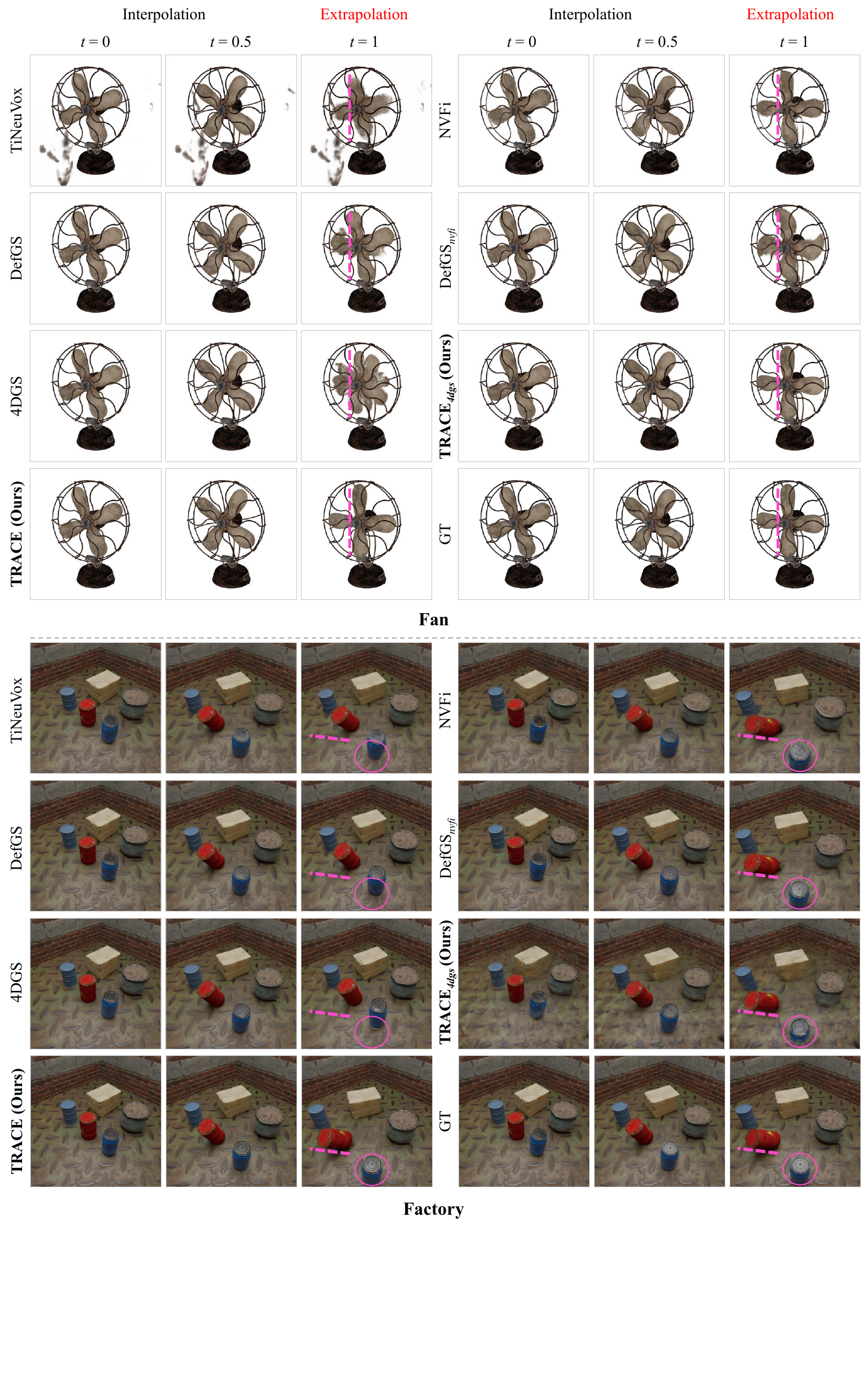}
    \vspace*{-25mm}
    \caption{Qualitative results of RGB view synthesis for interpolation and \textcolor{red}{extrapolation} tasks on Dynamic Object and Dynamic Indoor Scene datasets.}
    \label{fig:app_qual_res_indoor1}
\end{figure*}

\begin{figure*}[t]
    \centering
    \includegraphics[width=0.85\linewidth]{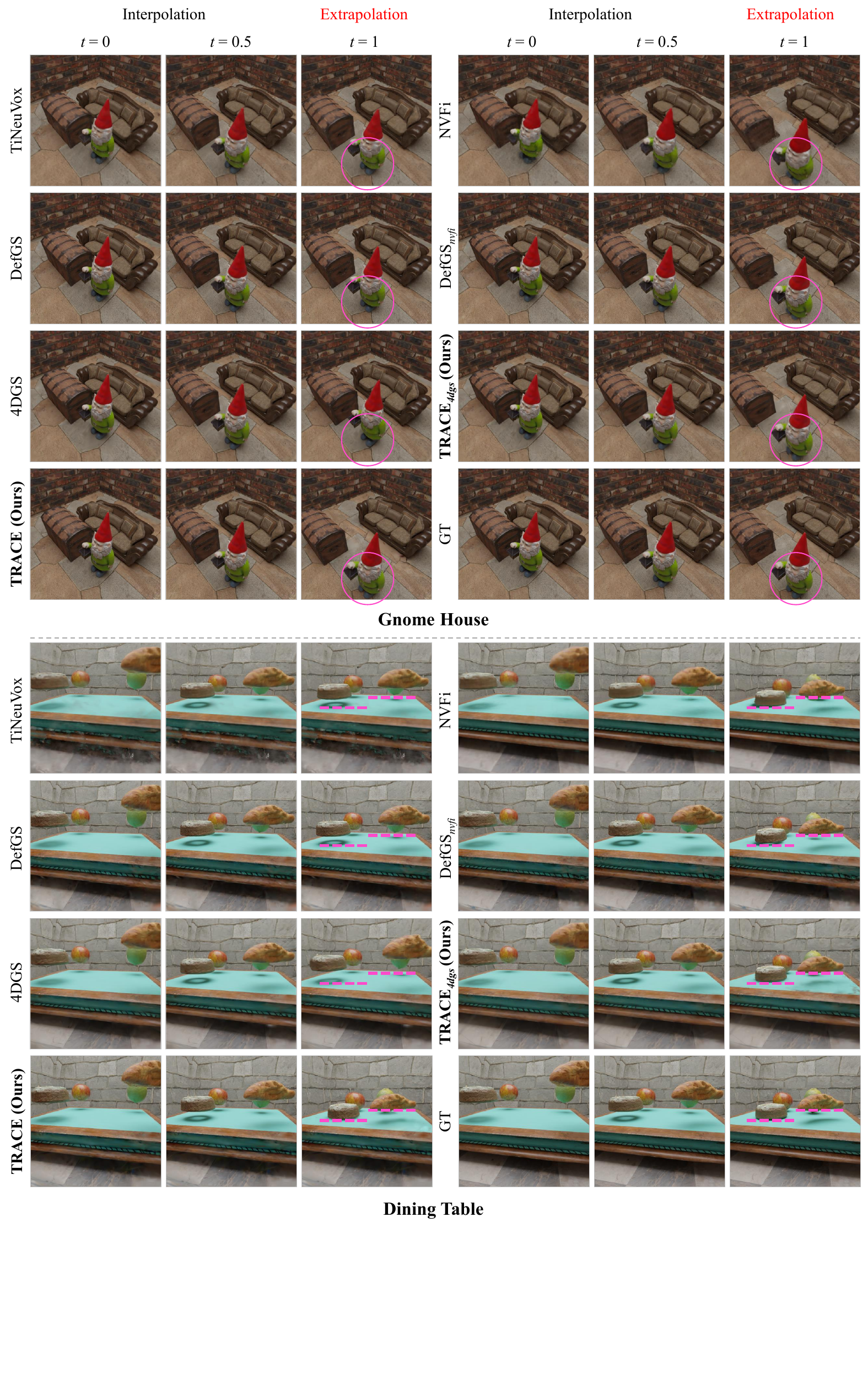}
    \vspace*{-25mm}
    \caption{Qualitative results of RGB view synthesis for interpolation and \textcolor{red}{extrapolation} tasks on Dynamic Indoor Scene dataset.}
    \label{fig:app_qual_res_indoor2}
\end{figure*}

Here we show the total per-scene results in Dynamic Indoor Scene Datasets in Table \ref{tab:indoor_all} and qualitative results in Figures \ref{fig:app_qual_res_indoor1}\&\ref{fig:app_qual_res_indoor2}, both for interpolation and extrapolation.

\begin{table*}[t]\tabcolsep=0.1cm 
\footnotesize
\centering
\caption{Per-scene quantitative results on Dynamic Indoor Scene dataset.}
\resizebox{1.0\linewidth}{!}{
\begin{tabular}{r|cccccc|cccccc}
\toprule
 & \multicolumn{6}{c|}{Gnome House} & \multicolumn{6}{c}{Chessboard} \\ \cmidrule{2-13} 
Methods & \multicolumn{3}{c}{Interpolation} & \multicolumn{3}{c|}{Extrapolation} & \multicolumn{3}{c}{Interpolation} & \multicolumn{3}{c}{Extrapolation} \\ \cmidrule{2-13}
 & PSNR$\uparrow$ & SSIM$\uparrow$ & LPIPS$\downarrow$ & PSNR$\uparrow$ & SSIM$\uparrow$ & LPIPS$\downarrow$ & PSNR$\uparrow$ & SSIM$\uparrow$ & LPIPS$\downarrow$ & PSNR$\uparrow$ & SSIM$\uparrow$ & LPIPS$\downarrow$ \\ \midrule
T-NeRF\citep{Pumarola2021} & 26.094 & 0.716 & 0.383 & 23.485 & 0.643 & 0.419 & 25.517 & 0.796 & 0.294 & 20.228 & 0.708 & 0.365 \\
D-NeRF\citep{Pumarola2021} & 27.000 & 0.745 & 0.319 & 21.714 & 0.641 & 0.367 & 24.852 & 0.774 & 0.308 & 19.455 & 0.675 & 0.384 \\
TiNeuVox\citep{Fang2022} & 30.646 & 0.831 & 0.253 & 21.418 & 0.699 & 0.326 & 33.001 & 0.917 & 0.177 & 19.718 & 0.765 & 0.310 \\
T-NeRF$_{PINN}$ & 15.008 & 0.375 & 0.668 & 16.200 & 0.409 & 0.651 & 16.549 & 0.457 & 0.621 & 17.197 & 0.472 & 0.618 \\
HexPlane$_{PINN}$ & 23.764 & 0.658 & 0.510 & 22.867 & 0.658 & 0.510 & 24.605 & 0.778 & 0.412 & 21.518 & 0.748 & 0.428 \\ 
NSFF\citep{Li2021c} & 31.418 & 0.821 & 0.294 & 25.892 & 0.750 & 0.327 & 32.514 & 0.810 & 0.201 & 21.501 & 0.805 & 0.282 \\ 
NVFi\citep{Li2023c}  & 30.667 & 0.824 & 0.277 & 30.408 & 0.826 & 0.273 & 30.394 & 0.888 & 0.215 & 27.840 & 0.872 & 0.219 \\
DefGS\citep{Yang2024} & 32.041 & 0.918 & 0.132 & 21.703 & 0.775 & 0.207 & 27.355 & 0.912 & 0.147 & 20.032 & 0.808 & 0.218 \\ 
DefGS$_{nvfi}$ & 32.881 & 0.919 & 0.132 & 33.630 & 0.953 & 0.077 & 26.200 & 0.907 & 0.156 & 26.730 & 0.917 & 0.110 \\ 
\midrule
\textbf{\nickname{}$_{4dgs}$ (Ours)} & 32.055 & 0.878 & 0.243 & 34.033 & 0.901 & 0.218 & 31.101 & 0.931 & 0.155 & 28.337 & 0.898 & 0.201 \\
\textbf{\nickname{} (Ours)} & \textbf{32.971} & \textbf{0.922} & \textbf{0.106} & \textbf{36.479} & \textbf{0.962} & \textbf{0.065} & \textbf{35.203} & \textbf{0.961} & \textbf{0.064} & \textbf{35.271} & \textbf{0.972} & \textbf{0.050} \\ 
\midrule

 & \multicolumn{6}{c|}{Factory} & \multicolumn{6}{c}{Dining Table} \\ \cmidrule{2-13} 
Methods & \multicolumn{3}{c}{Interpolation} & \multicolumn{3}{c|}{Extrapolation} & \multicolumn{3}{c}{Interpolation} & \multicolumn{3}{c}{Extrapolation} \\ \cmidrule{2-13}
 & PSNR$\uparrow$ & SSIM$\uparrow$ & LPIPS$\downarrow$ & PSNR$\uparrow$ & SSIM$\uparrow$ & LPIPS$\downarrow$ & PSNR$\uparrow$ & SSIM$\uparrow$ & LPIPS$\downarrow$ & PSNR$\uparrow$ & SSIM$\uparrow$ & LPIPS$\downarrow$ \\ \midrule
T-NeRF\cite{Pumarola2021} & 26.467 & 0.741 & 0.328 & 24.276 & 0.722 & 0.344 & 21.699 & 0.716 & 0.338 & 20.977 & 0.725 & 0.324 \\
D-NeRF\cite{Pumarola2021} & 28.818 & 0.818 & 0.252 & 22.959 & 0.746 & 0.303 & 20.851 & 0.725 & 0.319 & 19.035 & 0.705 & 0.341 \\
TiNeuVox\cite{Fang2022} & 32.684 & 0.909 & 0.148 & 22.622 & 0.810 & 0.229 & 23.596 & 0.798 & 0.274 & 20.357 & 0.804 & 0.258 \\
T-NeRF$_{PINN}$ & 16.634 & 0.446 & 0.624 & 17.546 & 0.480 & 0.609 & 16.807 & 0.486 & 0.640 & 18.215 & 0.548 & 0.595 \\
HexPlane$_{PINN}$ & 27.200 & 0.826 & 0.283 & 24.998 & 0.792 & 0.312 & 25.291 & 0.788 & 0.350 & 22.979 & 0.771 & 0.355 \\ 
NSFF\cite{Li2021c} & \textbf{33.975} & 0.919 & 0.152 & 26.647 & 0.855 & 0.196 & 19.552 & 0.665 & 0.464 & 22.612 & 0.770 & 0.351 \\ 
NVFi\cite{Li2023c}  & 32.460 & 0.912 & 0.151 & 31.719 & 0.908 & 0.154 & \textbf{29.179} & 0.885 & 0.199 & 29.011 & 0.898 & 0.171 \\ 
DefGS\cite{Yang2024} & 33.629 & \textbf{0.943} & 0.096 & 22.820 & 0.839 & 0.169 & 27.680 & 0.890 & 0.145 & 20.965 & 0.855 & 0.157 \\ 
DefGS$_{nvfi}$ & 33.643 & \textbf{0.943} & 0.097 & 33.049 & 0.954 & 0.062 & 27.957 & 0.891 & 0.145 & 30.975 & 0.955 & 0.060 \\ 
\midrule
\textbf{\nickname{}$_{4dgs}$ (Ours)} & 23.668 & 0.747 & 0.385 & 27.584 & 0.846 & 0.275 & 28.391 & \textbf{0.892} & 0.184 & 32.472 & 0.940 & 0.141 \\
\textbf{\nickname{} (Ours)} & 33.019 & \textbf{0.943} & \textbf{0.079} & \textbf{35.293} & \textbf{0.965} & \textbf{0.050} & 27.159 & 0.891 & \textbf{0.124} & \textbf{32.253} & \textbf{0.961} & \textbf{0.052} \\ \bottomrule
\end{tabular}
}
\label{tab:indoor_all}
\vspace{-0.1cm}
\end{table*}

\subsection{Additional Quantitative Results on NVIDIA Dynamic Scene Dataset}\label{sec:app_res_nvidia}
\begin{figure*}[t]
    \centering
    \includegraphics[width=0.85\linewidth]{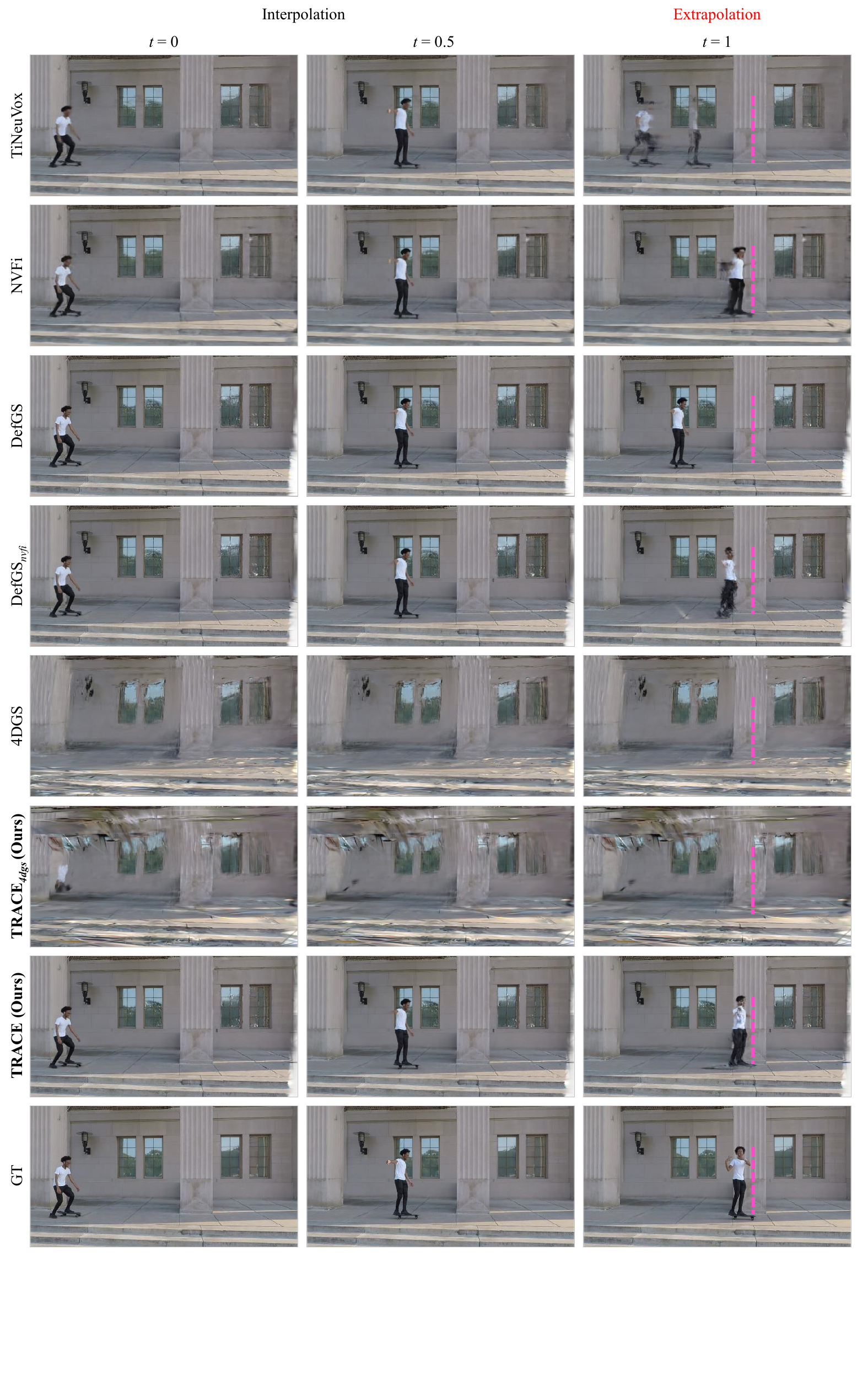}
    \vspace*{-20mm}
    \caption{Qualitative results of RGB view synthesis for interpolation and \textcolor{red}{extrapolation} tasks on ``Skating" scene of NVIDIA Dynamic Scene dataset.}
    \label{fig:app_qual_res_nvidia}
\end{figure*}

Here we show the total per-scene results on NVIDIA Dynamic Scene Dataset in Table \ref{tab:exp_extrapolation_real} and qualitative results in Figure \ref{fig:app_qual_res_nvidia}, both for interpolation and extrapolation.

\begin{table*}[t]\tabcolsep=0.1cm 
\centering
\caption{Quantitative results of our method and baselines on the NVIDIA Dynamic Scene dataset.}
\resizebox{1.\linewidth}{!}{
\begin{tabular}{r|cccccc|cccccc}
\toprule
\multirow{3}{*}{} & \multicolumn{6}{c|}{Truck} & \multicolumn{6}{c}{Skating} \\ \cmidrule{2-13} 
& \multicolumn{3}{c|}{Interpolation} & \multicolumn{3}{c|}{Extrapolation} & \multicolumn{3}{c|}{Interpolation} & \multicolumn{3}{c}{Extrapolation} \\ \cmidrule{2-13} 
& PSNR$\uparrow$ & SSIM$\uparrow$ & \multicolumn{1}{c|}{LPIPS$\downarrow$} & PSNR$\uparrow$ & SSIM$\uparrow$ & LPIPS$\downarrow$ & PSNR$\uparrow$ & SSIM$\uparrow$ & \multicolumn{1}{c|}{LPIPS$\downarrow$} & PSNR$\uparrow$ & SSIM$\uparrow$ & LPIPS$\downarrow$ \\ 
\cmidrule{2-13}
T-NeRF\citep{Pumarola2021} & 18.673 & 0.548 & \multicolumn{1}{c|}{0.447} & 18.176 & 0.567 & 0.447 & 27.483 & 0.820 & \multicolumn{1}{c|}{0.263} & 24.063 & 0.846 & 0.269 \\
D-NeRF\citep{Pumarola2021} & 17.660 & 0.554 & \multicolumn{1}{c|}{0.431} & 16.905 & 0.544 & 0.445 & 27.994 & 0.869 & \multicolumn{1}{c|}{0.187} & 24.361 & 0.873 & 0.208 \\
TiNeuVox\citep{Fang2022} & 27.230 & 0.846 & \multicolumn{1}{c|}{0.229} & 24.887 & 0.848 & 0.209 & 29.377 & 0.889 & \multicolumn{1}{c|}{0.202} & 24.224 & 0.878 & 0.220 \\
T-NeRF$_{PINN}$ & 15.241 & 0.413 & \multicolumn{1}{c|}{0.540} & 14.959 & 0.395 & 0.552 & 21.644 & 0.780 & \multicolumn{1}{c|}{0.338} & 20.990 & 0.814 & 0.303 \\
HexPlane$_{PINN}$ & 25.494 & 0.768 & \multicolumn{1}{c|}{0.337} & 24.991 & 0.768 & 0.325 & 24.447 & 0.867 & \multicolumn{1}{c|}{0.225} & 23.955 & 0.868 & 0.232 \\
NVFi\citep{Li2023c} & 27.276 & 0.840 & \multicolumn{1}{c|}{0.235} & 28.269 & 0.855 & 0.220 & \textbf{26.999} & 0.848 & \multicolumn{1}{c|}{0.227} & 28.654 & 0.896 & 0.208 \\
DefGS\citep{Yang2024} & \textbf{28.327} & \textbf{0.885} & \multicolumn{1}{c|}{0.115} & 24.947 & 0.875 & 0.131 & 24.997 & 0.900 & \multicolumn{1}{c|}{0.138} & 23.532 & 0.914 & 0.148 \\ 
DefGS$_{nvfi}$ & 28.169 & 0.884 & \multicolumn{1}{c|}{0.114} & 28.481 & \textbf{0.922} & 0.088 & 25.774 & 0.896 & \multicolumn{1}{c|}{0.141} & 26.577 & 0.931 & 0.115 \\ [+0.1em] 
\cmidrule{2-13} 
\textbf{\nickname{}$_{4dgs}$ (Ours)} & 19.330 & 0.435 & 0.551 & 21.073 & 0.610 & 0.481 & 17.905 & 0.428 & 0.554 & 24.034 & 0.831 & 0.298 \\
\textbf{\nickname{} (Ours)} & 28.070 & 0.878 & \multicolumn{1}{c|}{\textbf{0.107}} & \textbf{28.855} & 0.919 & \textbf{0.077} & 25.651 & \textbf{0.946} & \multicolumn{1}{c|}{\textbf{0.070}} & \textbf{29.827} & \textbf{0.946} & \textbf{0.070} \\[+0.1em] 
\bottomrule
\end{tabular}
}
\label{tab:exp_extrapolation_real}
\vspace{-0.1cm}
\end{table*}

\subsection{Additional Quantitative Results on Dynamic Multipart Dataset}\label{sec:app_res_multipart}
\begin{figure*}[t]
    \centering
    \includegraphics[width=0.85\linewidth]{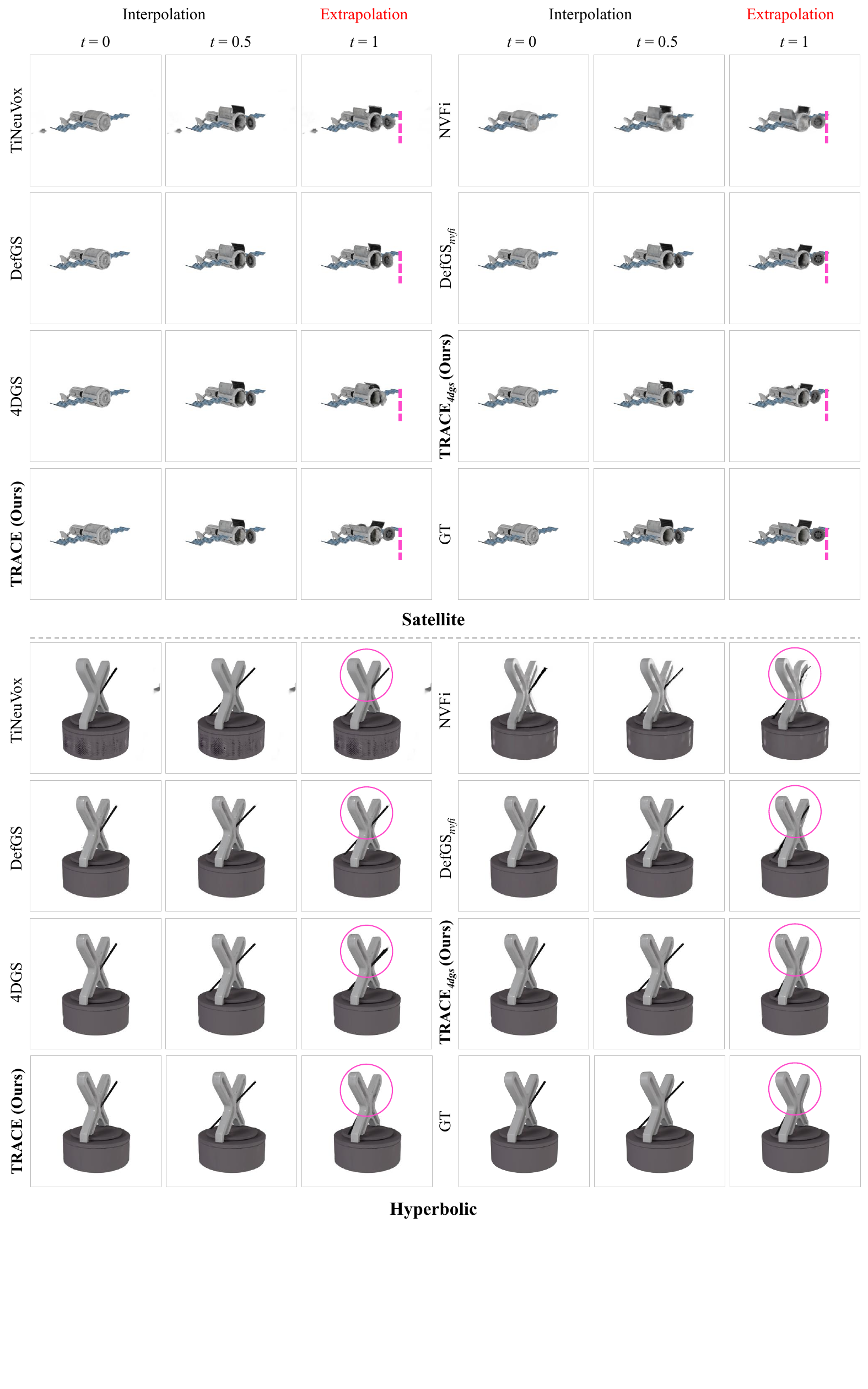}
    \vspace*{-25mm}
    \caption{Qualitative results of RGB view synthesis for interpolation and \textcolor{red}{extrapolation} tasks on Dynamic Multipart dataset.}
    \label{fig:app_qual_res_multipart}
\end{figure*}

Here we show the total per-scene results on our Dynamic Multipart Dataset in Table \ref{tab:multipart_all} and qualitative results in Figure \ref{fig:app_qual_res_multipart}, both for interpolation and extrapolation.

\begin{table*}[ht]\tabcolsep=0.1cm 
\footnotesize
\centering
\caption{Per-scene quantitative results on Dynamic Multipart dataset.}
\resizebox{1.0\linewidth}{!}{
\begin{tabular}{r|cccccc|cccccc}
\toprule
& \multicolumn{6}{c|}{Folding Chair} & \multicolumn{6}{c}{Hyperbolic Slot} \\ \cmidrule{2-13} 
Methods & \multicolumn{3}{c}{Interpolation} & \multicolumn{3}{c|}{Extrapolation} & \multicolumn{3}{c}{Interpolation} & \multicolumn{3}{c}{Extrapolation} \\ \cmidrule{2-13}
& PSNR$\uparrow$ & SSIM$\uparrow$ & LPIPS$\downarrow$ & PSNR$\uparrow$ & SSIM$\uparrow$ & LPIPS$\downarrow$ & PSNR$\uparrow$ & SSIM$\uparrow$ & LPIPS$\downarrow$ & PSNR$\uparrow$ & SSIM$\uparrow$ & LPIPS$\downarrow$ \\ \midrule
T-NeRF\citep{Pumarola2021} & 10.146 & 0.598 & 0.537 & 10.260 & 0.586 & 0.548 & 7.437 & 0.424 & 0.749 & 7.098 & 0.404 & 0.739 \\
D-NeRF \citep{Pumarola2021} & 11.681 & 0.717 & 0.437 & 13.177 & 0.765 & 0.357 & 7.279 & 0.485 & 0.714 & 7.547 & 0.468 & 0.695 \\
TiNeuVox\citep{Fang2022} & 34.160 & 0.984 & 0.039 & 13.391 & 0.808 & 0.199 & 28.637 & 0.955 & 0.083 & 25.436 & 0.973 & 0.040 \\
NVFi\citep{Li2023c}  & 27.748 & 0.962 & 0.049 & 23.433 & 0.940 & 0.063 & 25.487 & 0.944 & 0.057 & 25.757 & 0.956 & 0.039 \\
DefGS\citep{Yang2024} & 37.319 & 0.995 & 0.009 & 13.682 & 0.820 & 0.169 & 31.780 & 0.983 & 0.030 & 25.631 & 0.981 & 0.020 \\ 
DefGS$_{nvfi}$ & 37.269 & 0.994 & 0.009 & 25.404 & 0.962 & 0.022 & 32.506 & 0.985 & 0.025 & 29.351 & 0.988 & 0.012 \\ 
\midrule
\textbf{\nickname{}$_{4dgs}$ (Ours)} & 36.707 & 0.993 & 0.012 & 26.993 & 0.967 & 0.033 & \textbf{33.507} & \textbf{0.987} & 0.020 & \textbf{37.502} & 0.994 & 0.008 \\
\textbf{\nickname{} (Ours)} & \textbf{40.031} & \textbf{0.996} & \textbf{0.005} & \textbf{29.588} & \textbf{0.982} & \textbf{0.010} & 33.041 & 0.986 & \textbf{0.017} & 36.437 & \textbf{0.995} & \textbf{0.005} \\
\midrule

& \multicolumn{6}{c|}{Satellite} & \multicolumn{6}{c}{Stove} \\ \cmidrule{2-13}
Methods & \multicolumn{3}{c}{Interpolation} & \multicolumn{3}{c|}{Extrapolation} & \multicolumn{3}{c}{Interpolation} & \multicolumn{3}{c}{Extrapolation} \\ \cmidrule{2-13}
& PSNR$\uparrow$ & SSIM$\uparrow$ & LPIPS$\downarrow$ & PSNR$\uparrow$ & SSIM$\uparrow$ & LPIPS$\downarrow$ & PSNR$\uparrow$ & SSIM$\uparrow$ & LPIPS$\downarrow$ & PSNR$\uparrow$ & SSIM$\uparrow$ & LPIPS$\downarrow$ \\ \midrule
T-NeRF\citep{Pumarola2021} & 14.614 & 0.754 & 0.307 & 14.468 & 0.751 & 0.328 & 7.134 & 0.490 & 0.605 & 8.429 & 0.562 & 0.531 \\
D-NeRF\citep{Pumarola2021} & 17.991 & 0.930 & 0.100 & 17.252 & 0.926 & 0.102 & 16.165 & 0.856 & 0.262 & 15.400 & 0.908 & 0.205 \\
TiNeuVox\citep{Fang2022} & 33.061 & 0.983 & 0.035 & 28.627 & 0.978 & 0.032 & 23.969 & 0.943 & 0.109 & 15.760 & 0.934 & 0.087 \\
NVFi\citep{Li2023c}  & 29.644 & 0.973 & 0.029 & 30.075 & 0.975 & 0.027 & 27.186 & 0.959 & 0.072 & 21.675 & 0.950 & 0.054 \\
DefGS\citep{Yang2024} & 36.832 & \textbf{0.993} & 0.007 & 27.622 & 0.979 & 0.016 & 32.607 & 0.989 & 0.029 & 15.721 & 0.941 & 0.063 \\ 
DefGS$_{nvfi}$ & 36.640 & \textbf{0.993} & \textbf{0.007} & \textbf{34.282} & \textbf{0.990} & \textbf{0.007} & 21.134 & 0.988 & 0.029 & 24.781 & 0.977 & 0.027  \\ 
\midrule
\textbf{\nickname{}$_{4dgs}$ (Ours)} & 35.722 & 0.992 & 0.010 & 30.804 & 0.986 & 0.012 & \textbf{38.037} & \textbf{0.992} & 0.023 & 33.969 & 0.991 & 0.012 \\
\textbf{\nickname{} (Ours)} & \textbf{36.938} & \textbf{0.993} & \textbf{0.007} & 32.093 & 0.988 & \textbf{0.007} & 33.062 & 0.990 & \textbf{0.015} & \textbf{35.805} & \textbf{0.994} & \textbf{0.007} \\
\bottomrule
\end{tabular}
}
\label{tab:multipart_all}
\vspace{-0.1cm}
\end{table*}

\subsection{Additional qualitative results for extrapolation beyond dataset time spans}\label{sec:app_longer_extrap}
\begin{figure*}[t]
    \centering
    \includegraphics[width=0.8\linewidth]{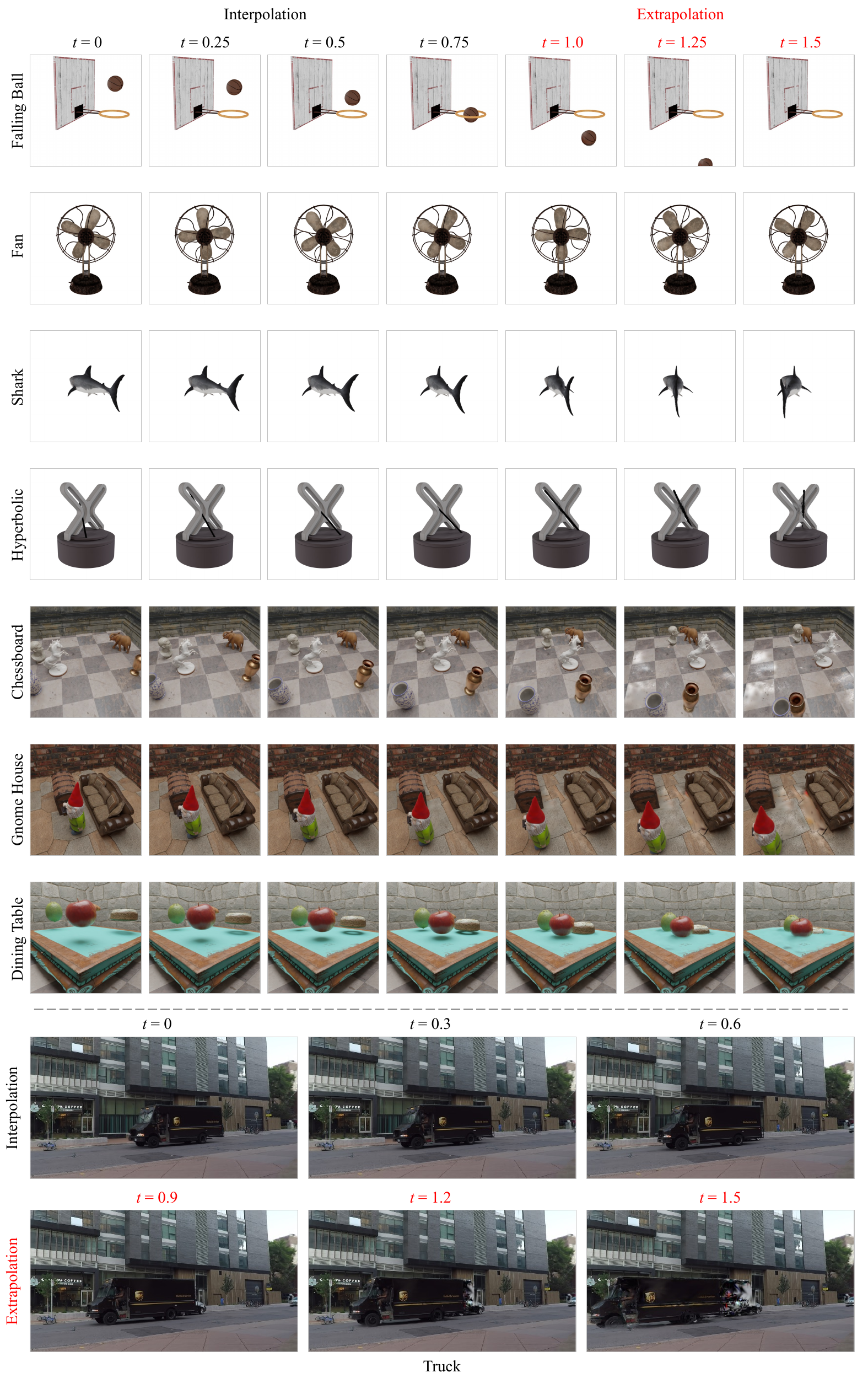}
    \caption{Qualitative results of RGB view synthesis for longer extrapolation from our method.}
    \label{fig:app_longer_extrap}
\end{figure*}

We list some meaningful longer extrapolation results from each dataset here in Figure \ref{fig:app_longer_extrap}. In our dataset, the training period lasts from $t=0$ to $t=0.75$ and the extrapolation period lasts from $t=0.75$ to $t=1.0$. Here we show the qualitative results till $t=1.5$, which is already twice the training period. We can see that our method can still obtain physically meaningful future frame prediction in particularly high quality.

\subsection{Additional Qualitative Results for Object/Part Segmentation}\label{sec:app_res_seg_point}
\begin{figure*}[t]
    \centering
    \includegraphics[width=0.75\linewidth]{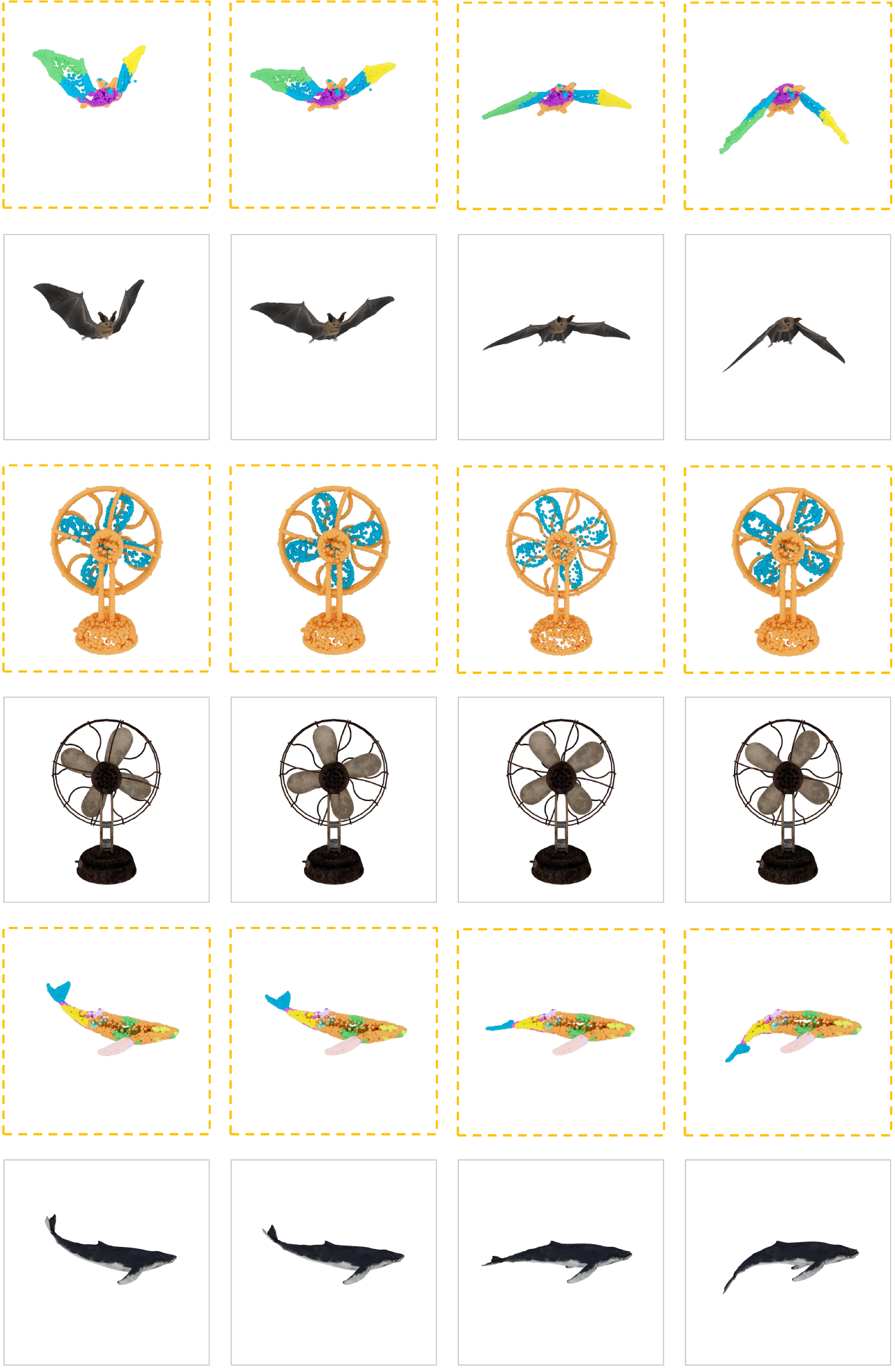}
    \caption{Qualitative results for Object/Part Segmentation on Dynamic Object dataset.}
    \label{fig:app_seg_obj}
\end{figure*}

\begin{figure*}[t]
    \centering
    \includegraphics[width=0.75\linewidth]{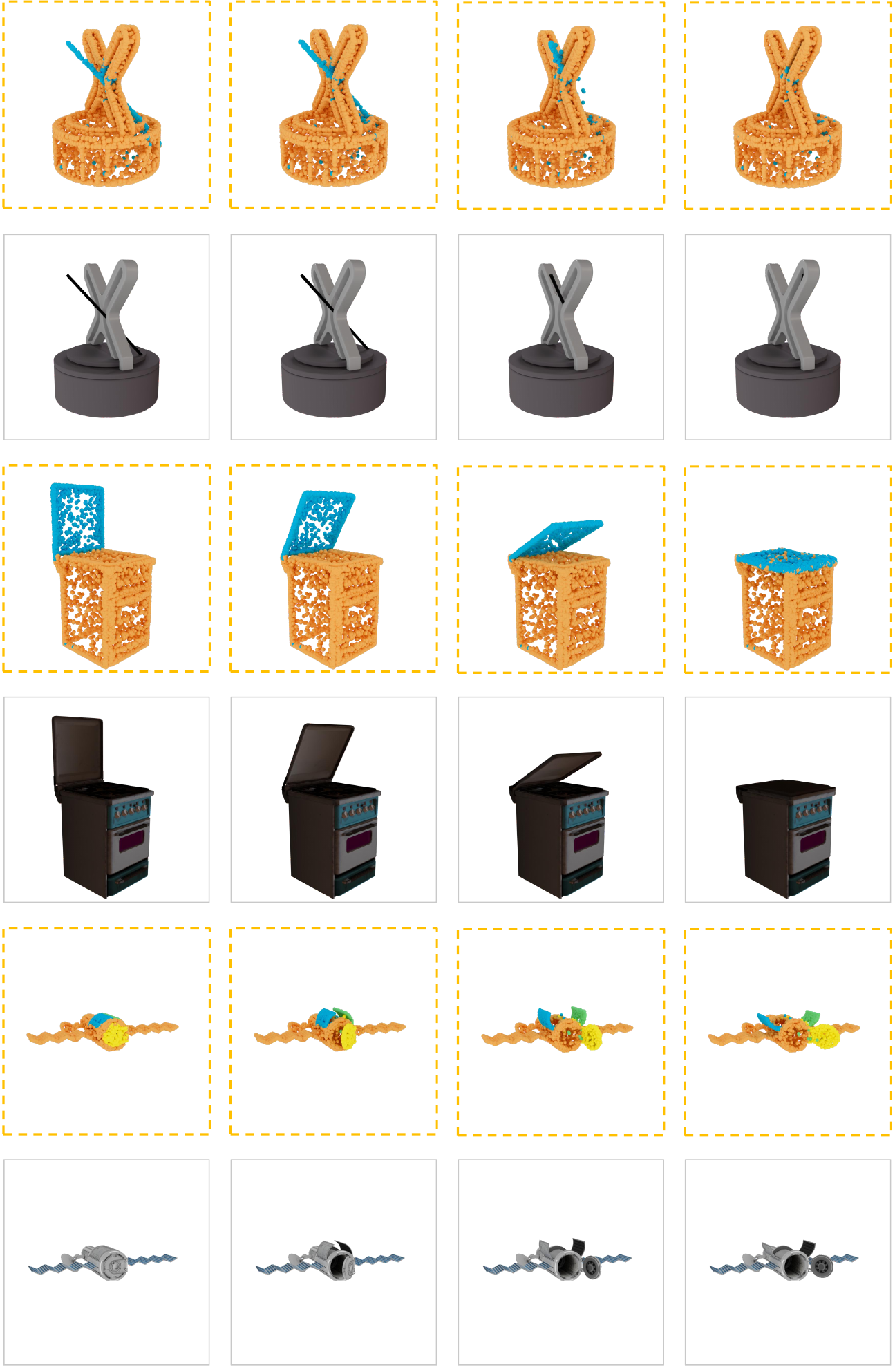}
    \caption{Qualitative results for Object/Part Segmentation on Dynamic Multipart dataset.}
    \label{fig:app_seg_newobj}
\end{figure*}

\begin{figure*}[t]
    \centering
    \includegraphics[width=0.85\linewidth]{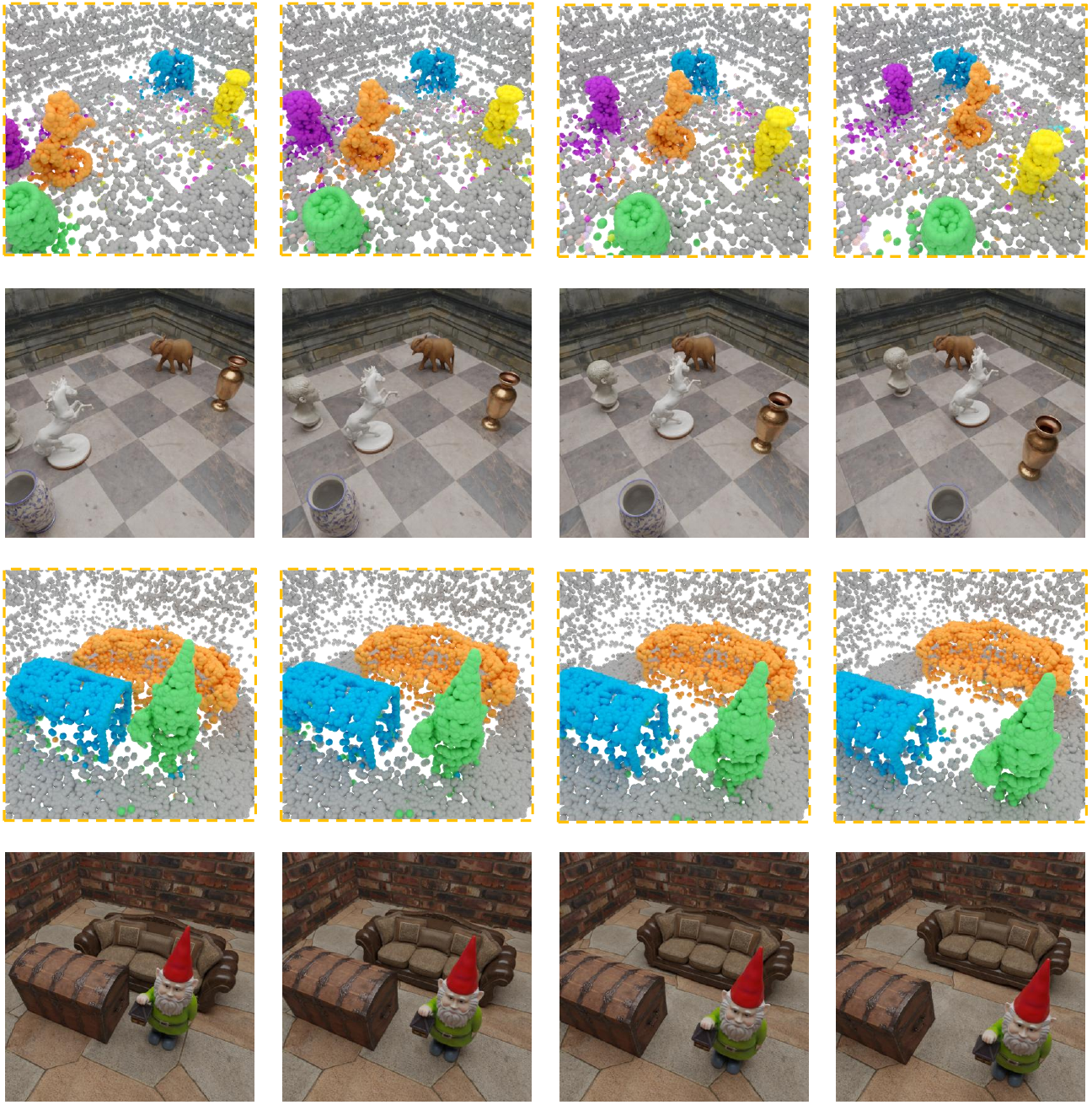}
    \caption{Qualitative results for Object/Part Segmentation on Dynamic Indoor Scene dataset.}
    \label{fig:app_seg_indoor}
\end{figure*}

Figures \ref{fig:app_seg_obj}, \ref{fig:app_seg_newobj} and \ref{fig:app_seg_indoor} show more qualitative results for the autonomous object or part segmentation based on the learned physical parameters via the simple K-means clustering algorithm. 

\subsection{Additional Qualitative Results for Segmentation on Dynamic Indoor Scene Dataset}\label{sec:app_res_seg_mask}
\begin{figure*}[t]
    \centering
    \includegraphics[width=0.9\linewidth]{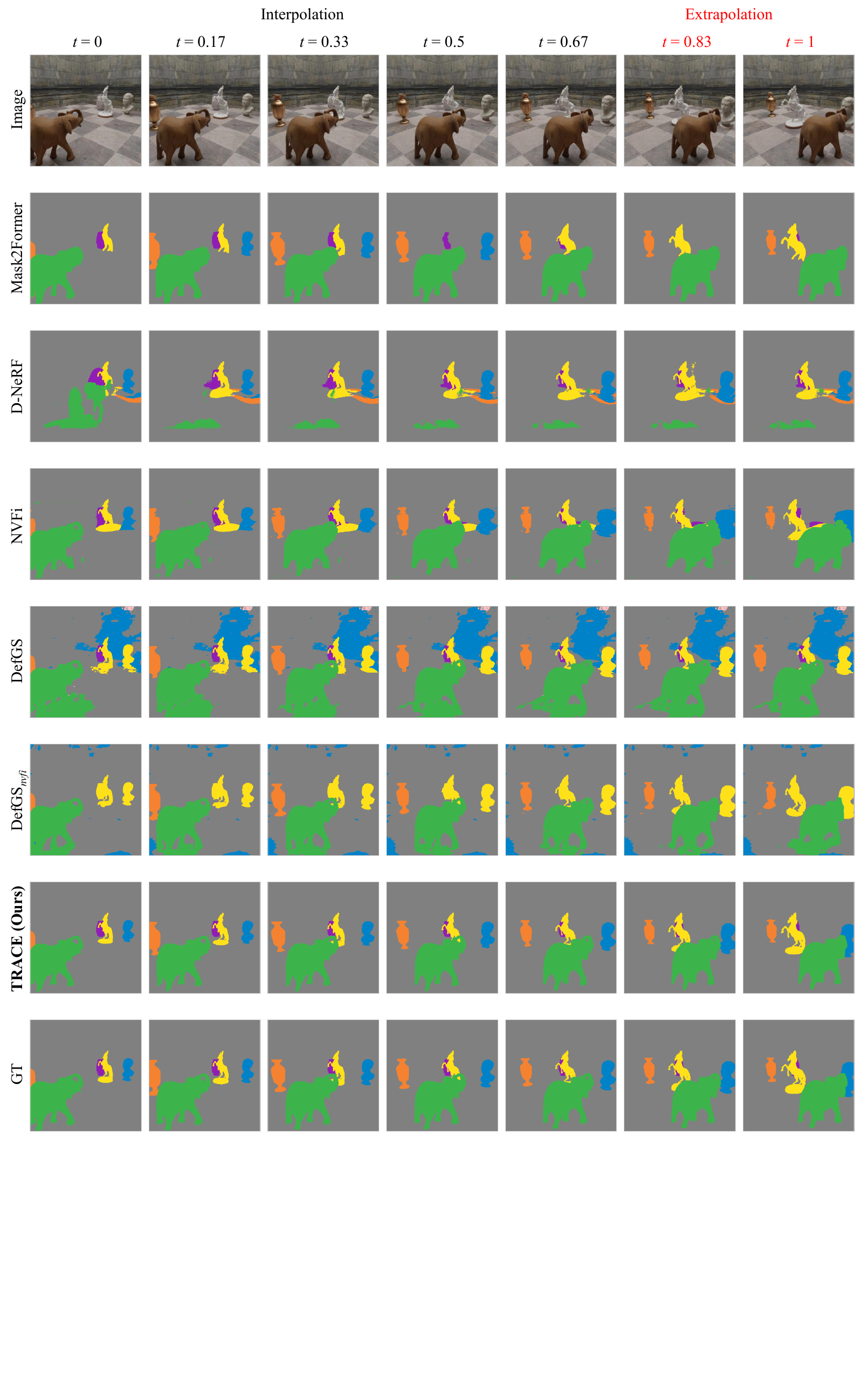}
    \vspace*{-40mm}
    \caption{Qualitative results for object segmentation on ``Chessboard" of Dynamic Indoor Scene dataset.}
    \label{fig:app_qual_segm1}
\end{figure*}

\begin{figure*}[t]
    \centering
    \includegraphics[width=0.9\linewidth]{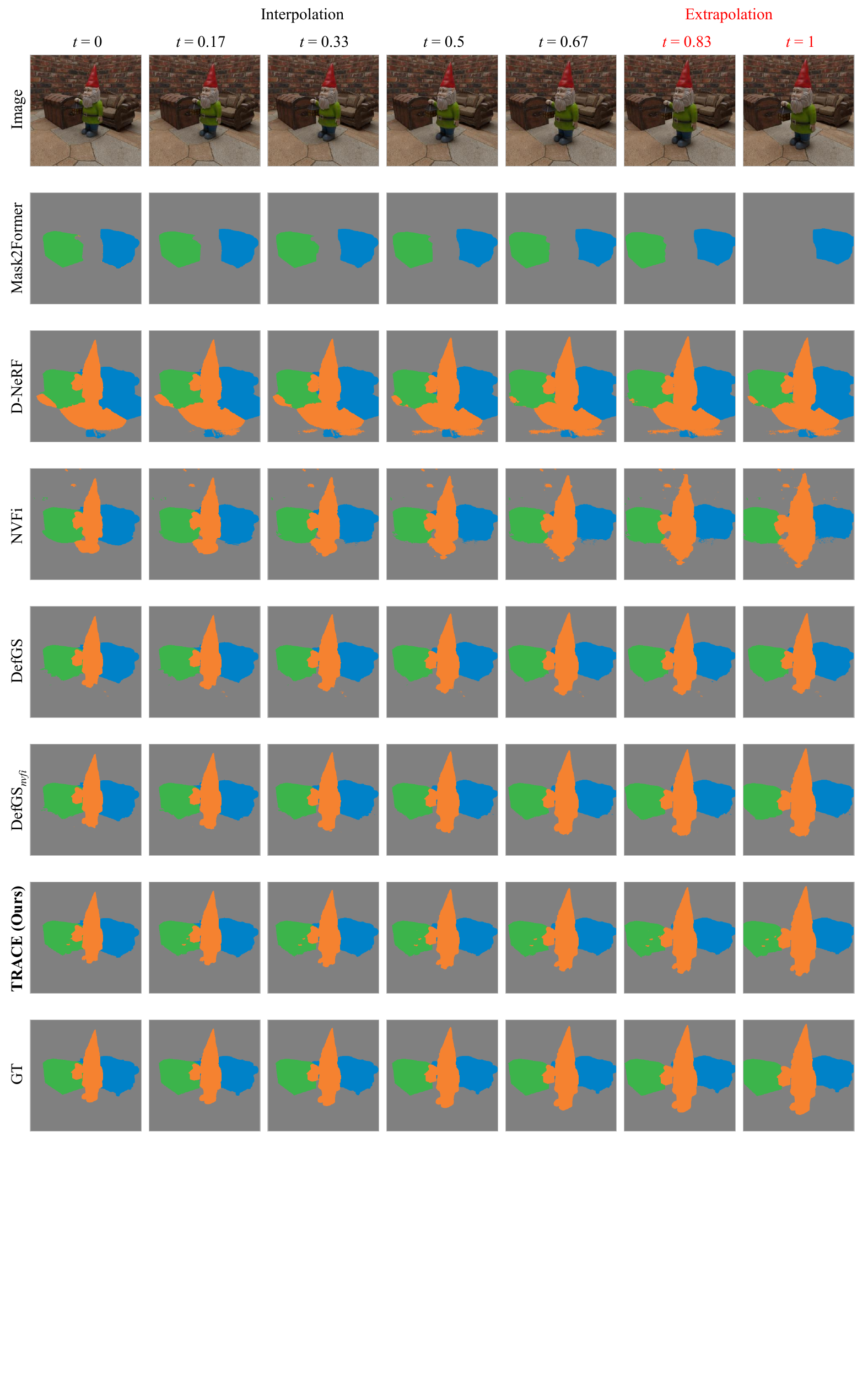}
    \vspace*{-40mm}
    \caption{Qualitative results for object segmentation on ``Gnome House" of Dynamic Indoor Scene dataset.}
    \label{fig:app_qual_segm2}
\end{figure*}

\begin{figure*}[t]
    \centering
    \includegraphics[width=0.9\linewidth]{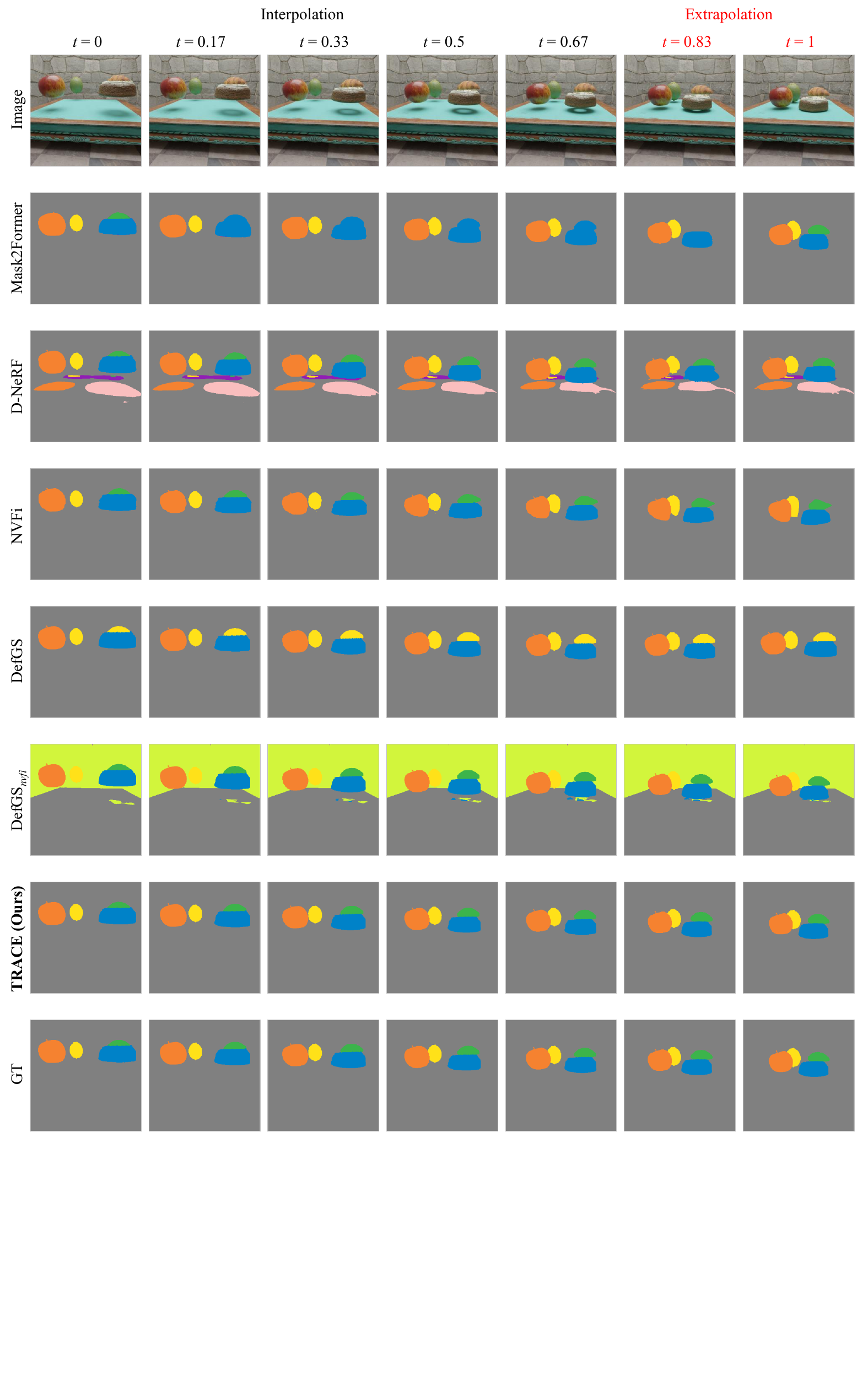}
    \vspace*{-40mm}
    \caption{Qualitative results for object segmentation on ``Dining Table" of Dynamic Indoor Scene dataset.}
    \label{fig:app_qual_segm3}
\end{figure*}

\begin{figure*}[t]
    \centering
    \includegraphics[width=0.9\linewidth]{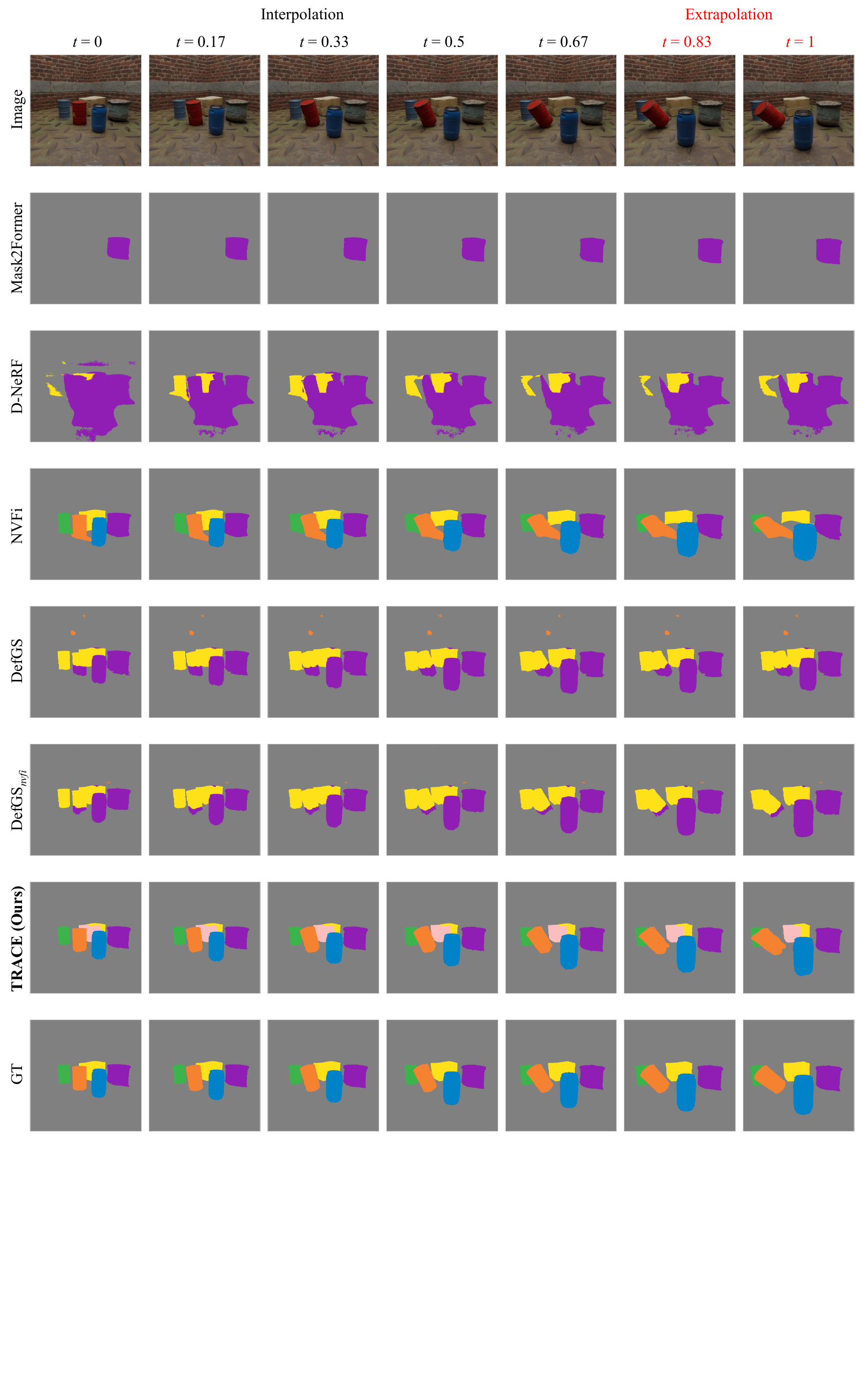}
    \vspace*{-40mm}
    \caption{Qualitative results for object segmentation on ``Factory" of Dynamic Indoor Scene dataset.}
    \label{fig:app_qual_segm4}
\end{figure*}

Figures \ref{fig:app_qual_segm1}, \ref{fig:app_qual_segm2}, \ref{fig:app_qual_segm3}, \&\ref{fig:app_qual_segm4} shows qualitative results for the rendered mask on Dynamic Indoor Scene dataset.

\end{document}